\documentclass[runningheads]{llncs}

\usepackage[utf8]{inputenc}
\usepackage{listings}
\usepackage{protobuf/lang}  
\usepackage{protobuf/style} 
\usepackage{listings}
\usepackage{graphicx}
\usepackage{multirow}
\usepackage{enumitem}
\usepackage{hyperref}
\usepackage{subcaption}

\usepackage{todonotes}

\usepackage{geometry}
\title{EvoCraft: A New Challenge for Open-Endedness}

\author{Djordje Grbic$^{1}$, Rasmus Berg Palm$^{1}$, Elias Najarro$^{1}$, Claire Glanois$^{2}$, Sebastian Risi$^{1, 3}$}

\authorrunning{D. Grbic et al.}

\titlerunning{EvoCraft: A New Challenge for Open-Endedness}

\institute{$^{1}$IT University of Copenhagen, $^{2}$Shanghai University,  $^{3}$modl.ai}

\begin{document}

\maketitle

\begin{abstract}
    This paper introduces EvoCraft, a framework for Minecraft designed to study open-ended algorithms. We introduce an  API that provides an open-source Python interface for communicating with Minecraft to place and track blocks. In contrast to previous work in Minecraft that focused on learning to play the game, the grand challenge we pose here is to automatically search for increasingly complex artifacts in an open-ended fashion. Compared to other environments used to study open-endedness, Minecraft allows the construction of almost any kind of structure, including actuated machines with circuits and mechanical components. 
    We present initial baseline results in evolving simple Minecraft creations through both interactive and automated evolution. While evolution succeeds when tasked to grow a structure towards a specific target, it is unable to find a solution 
   when rewarded for creating a simple machine that moves. Thus, EvoCraft offers a challenging new environment for automated search methods (such as evolution) to find complex artifacts that we hope will spur the development of more open-ended algorithms. A Python implementation of the EvoCraft framework is available at: \href{https://github.com/real-itu/Evocraft-py}{github.com/real-itu/Evocraft-py}.
\end{abstract}

\section{Introduction}
Artificial intelligence (AI) approaches have shown remarkable advances in the last couple of years, solving increasingly complex challenges. A key driver in these advances has been specific environments and competitions that allowed different approaches to be easily compared. 
For example, developing frameworks and learning environments, such as the StarCraft II Learning Environment \cite{vinyals2017starcraft} has spurred the development of many recent advances in the field \cite{jaderberg2017population,vinyals2017starcraft}.

A grand research challenge with relatively less progress has been the goal of open-endedness \cite{soros2014identifying,packard2019overview,stanley2017open}. Similarly to how evolution in nature created the seemingly endless array of novel and adaptive forms, the goal in open-endedness is to create similar open-ended algorithms in artificial life simulations \cite{bedau2000open}. 

\begin{figure}
     \centering
    \includegraphics[width=\textwidth]{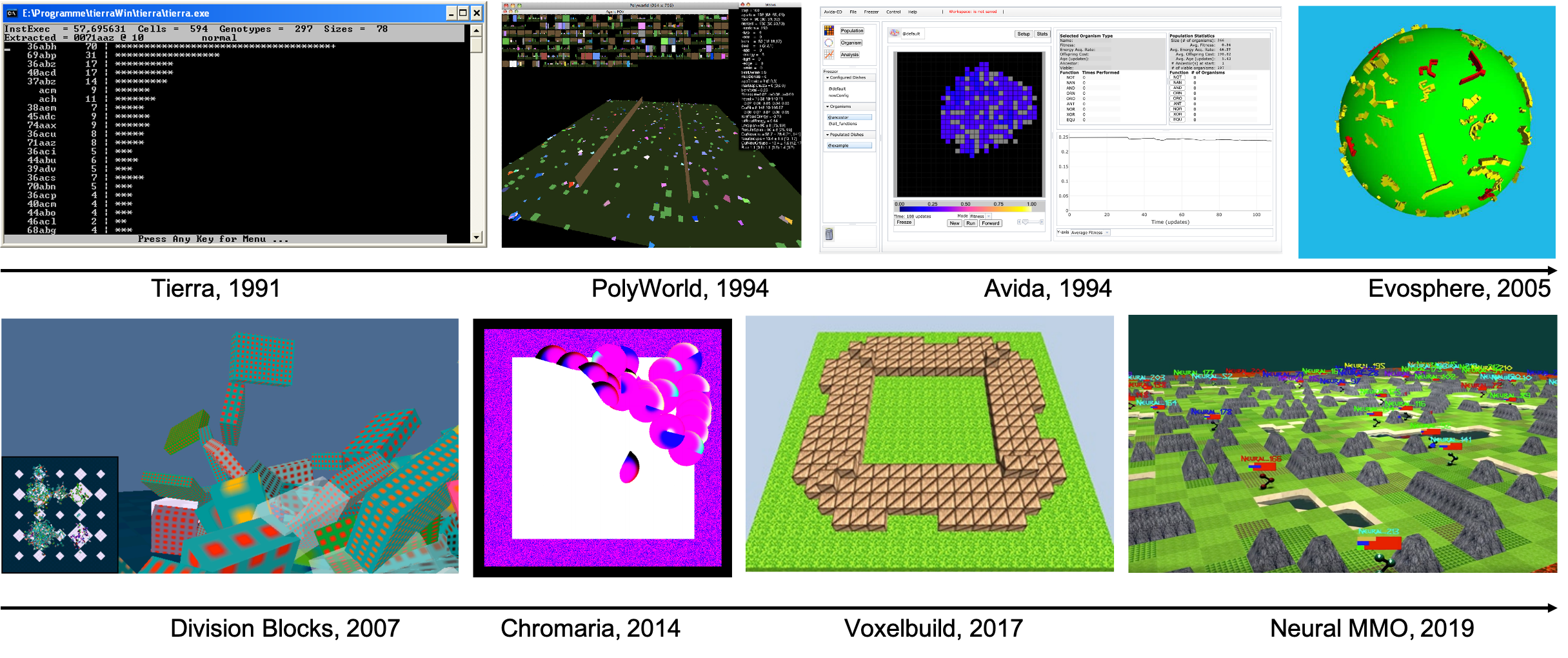}
     \caption{A selection of existing OOE frameworks.}
     \label{fig:alife_worlds}
\end{figure}

\begin{figure}
     \centering
     \begin{subfigure}[b]{0.48\textwidth}
         \centering
         \includegraphics[width=\textwidth]{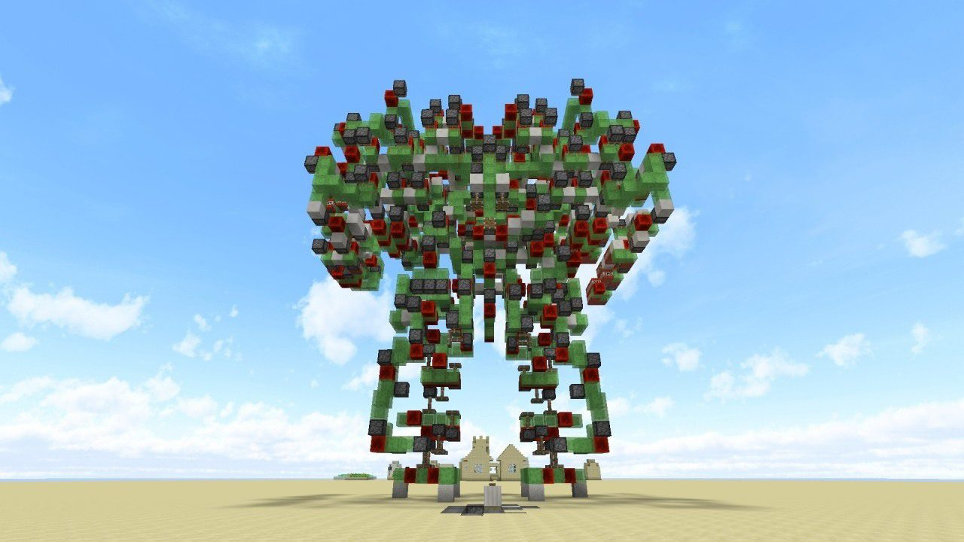}
     \end{subfigure}
     \begin{subfigure}[b]{0.48\textwidth}
         \centering
         \includegraphics[width=\textwidth]{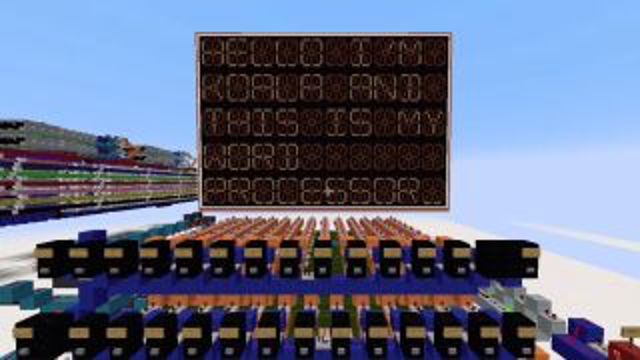}
     \end{subfigure}
     \begin{subfigure}[b]{0.48\textwidth}
         \centering
         \includegraphics[width=\textwidth]{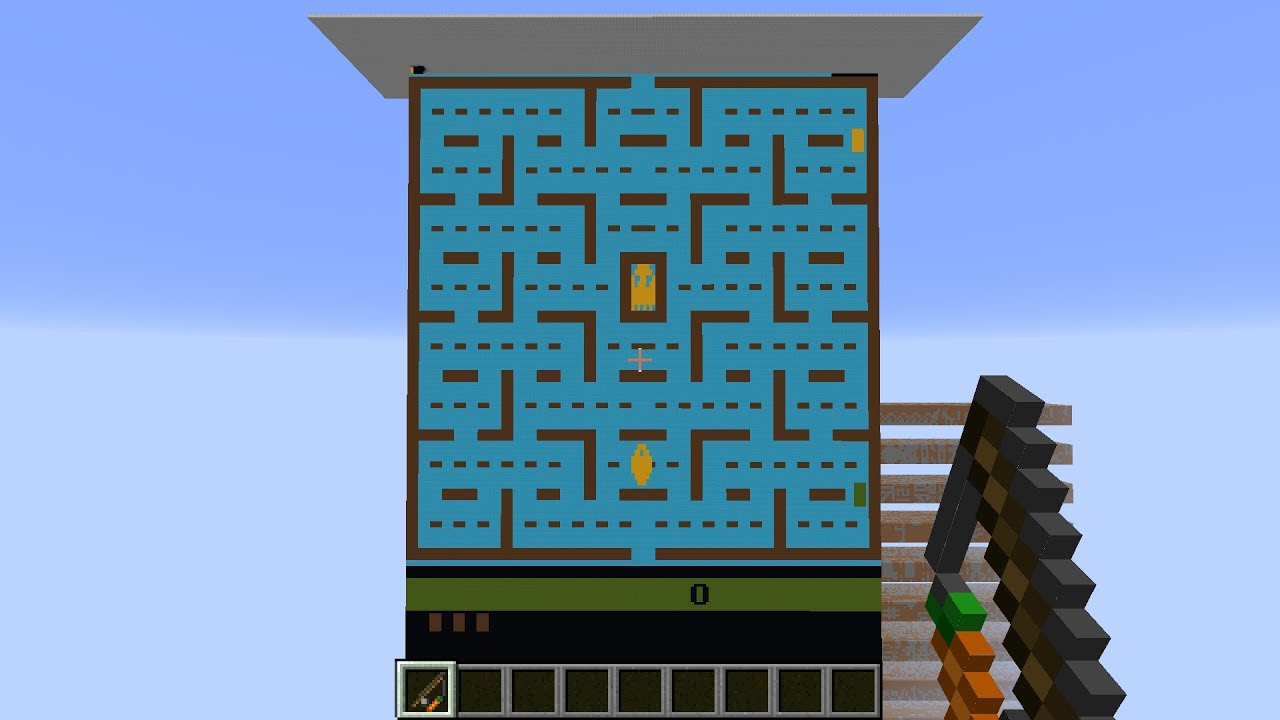}
     \end{subfigure}
      \begin{subfigure}[b]{0.48\textwidth}
         \centering
         \includegraphics[width=\textwidth]{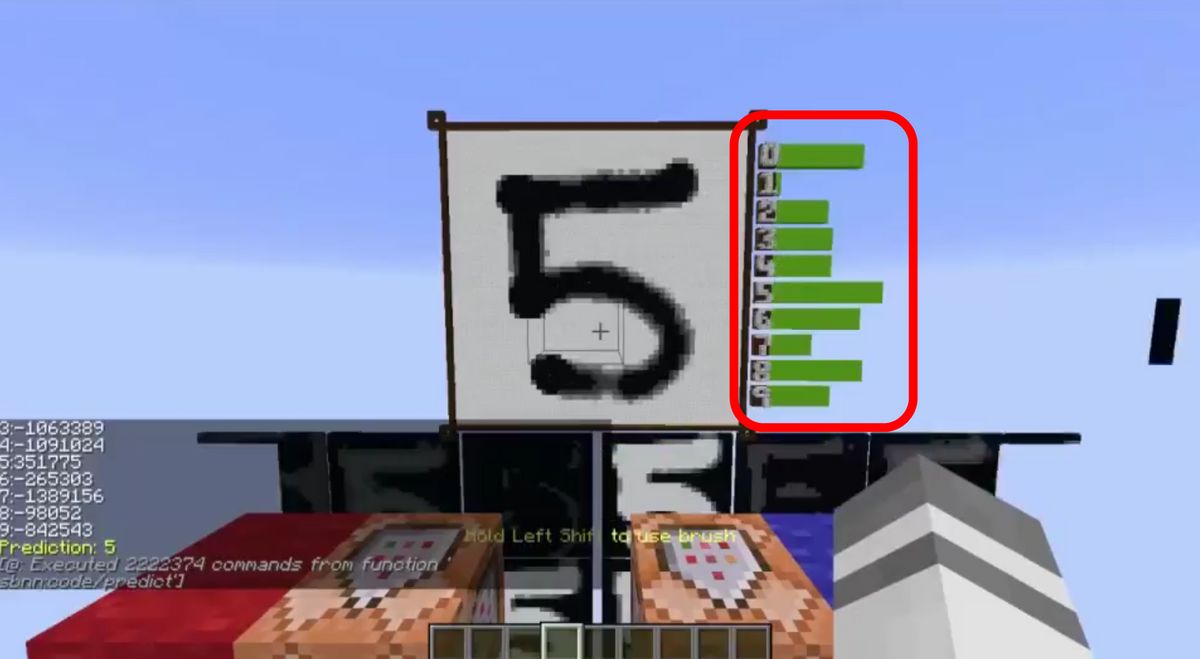}
     \end{subfigure}
     \caption{Examples of human-build structures in Minecraft: A large moving robot \cite{minecraft_robot}, a functioning word processors \cite{wordprocessor}, Atari 2600 emulator \cite{emulatorAtari} and neural network digit classifier  \cite{digitClassifier}.}
     \label{fig:minecraft_creations}
\end{figure}

However, while current alife environments have produced a large variety of important evolutionary insights  \cite{ray1991approach,ofria2004avida,miconi2005virtual,yaeger1994computational,soros2014identifying,pugh2017major}, 
they are often still limited in the type of open-ended evolutionary dynamics that can emerge. 
Some of these alife environments are shown in Figure~\ref{fig:alife_worlds}. Compared to structures players were able to built in Minecraft (Figure~\ref{fig:minecraft_creations}), current alife environments are somewhat limited in the type of behaviors or artifacts that  can be discovered. Additionally, the myriad of different alife worlds that often investigate different alife  aspects can make it difficult to compare approaches to each other.

In this paper, we propose to use Minecraft as the perfect environment for the study of artificial life and open-endedness in particular. Minecraft is a voxel-based environment in which the basic building blocks are different types of blocks such as wood, stone, glass, water, etc.  
Especially the addition of ``redstone'' circuit components in Minecraft (i.e.\ blocks that support circuits and mechanical components), has allowed players to build amazing structures, such as moving robots, fully functioning word processors, or even Atari 2600 emulators (Figure~\ref{fig:minecraft_creations}). 
Compared to most alife domains, a major benefit of Minecraft is that everything can be built from a finite set of different and simple buildings blocks, which aligns well with the machinery of biological systems that only use a few building blocks to synthesizing the multitude of complex chemicals that form the basis of all organic life.
 
In this paper, we introduce the API, framework, and first results for our EvoCraft environment. EvoCraft  is implemented as a mod for Minecraft that allows clients to manipulate blocks in a running Minecraft server programmatically through an API. The framework is specifically developed to facilitate experiments in artificial evolution and other optimization algorithms. We show that evolving simple artefacts, such as structures that grow towards a particular target is straightforward with our developed API. We also show that evolving a Minecraft machine that moves is a challenging problem, for which traditional fitness-based approaches fail and make little progress. We therefore argue that EvoCraft offers an exciting and challenging environment for experiments in artificial life and open-ended evolution. In addition to releasing this API, we will run EvoCraft competitions at different evolutionary and artificial life conferences, starting with GECCO 2021.

One of the ideas of open-endedness is to be surprised by its creations. While previous domains were limited in the type of structures that could be evolved, Minecraft could offer a domain in which an open-ended discovery engine could generate truly surprising artifacts.

\section{Related work}
This section reviews work on environments used to study open-ended evolution and previous work that uses Minecraft to test different AI approaches. 

\subsection{Open-Endedness Frameworks}
Researchers have explored and developed a variety of different environments to study open-endedness \cite{bedau1998classification,bohm2017mabe,chan2018lenia,harrington2019escalation,miconi2005virtual,ofria2004avida,pugh2017major,ray1991approach,soros2014identifying,spector2007division,suarez2019neural,taylor1999artificial,yaeger1994computational}, some of which are shown in Figure~\ref{fig:alife_worlds}. One of the first examples of trying to achieve open-ended evolution in an artificial world was Tierra \cite{ray1991approach}. In Tierra programs that compete for CPU cycles evolve, being composed of machine code instructions. Avida \cite{ofria2004avida} is another artificial life world, which was inspired by Tierra, but here creatures are rewarded for their ability to perform various computations. Experiments in these artificial life environments can lead to important insights into evolutionary dynamics. For example, in a landmark study Lenski et al. \cite{lenski2003evolutionary} showed how complex behaviors can evolve by building on simpler functions with digital organisms in Avida. 

Other examples of such alife worlds include Geb \cite{bedau1998classification}, Evosphere \cite{miconi2005virtual} and PolyWorld \cite{yaeger1994computational}, in which populations of creatures evolve in 3D and 2D worlds, learning to interact and fight other creatures. Another interesting example is Division Blocks \cite{spector2007division}, in which 3D creatures made out of blocks evolve to grow, shrink, exchange resources, create joints, and actuate joints.  

More recently, researchers have studied in more detail what might be the necessary ingredients for open-ended dynamics to emerge. Within an alife world called Chromaria, Soros and Stanley \cite{soros2014identifying}, identified conditions such as that each individual must satisfy some minimum criteria before it can reproduce or that individuals should be able to decide where and how to interact with their world. If these conditions will be similar or different within the proposed EvoCraft environment is an interesting future research direction. 

Recent work related to EvoCraft is the  
Voxelbuild virtual sandbox environment  \cite{pugh2017major} (Figure~\ref{fig:alife_worlds}). In this environment, which is inspired by Minecraft, agents are evolving to build  complex block structures. The environment was designed to test an important facet of open-ended evolution, which are the mechanisms behind major evolutionary transitions. The authors observe that in some evolutionary runs, these transitions occur, such as the ability for the agents to figure out how to place blocks vertically. This work shows the potential for a virtual sandbox environment to study evolution and serves as a good proof of concept for the more complex EvoCraft environment introduced in this paper.     


A more recent approach, inspired by earlier multi-agent alife environments \cite{bedau1998classification,miconi2005virtual,yaeger1994computational}, is 
Neural MMO \cite{suarez2019neural}. The Neural MMO is a multi-agent system that features a large population of agents. In contrast to previously introduced multi-agent alife environments, Neural MMO is focused on training a large number of agents through reinforcement learning approaches to forage and to engage other agents in combat. Similar to the Neural MMO, Minecraft naturally supports a large number of agents and additionally also features the ability to build structures. 

While the environments presented in this section are undoubtedly useful and led to groundbreaking discoveries \cite{lenski2003evolutionary}, we argue here that Minecraft offers a naturally next step to  study open-endedness on a larger scale.  

In addition to developing different alife environments to foster open-ended evolution, recently interest has increased in creating algorithms for more open-ended learning. For example, a class of algorithms called quality diversity (QD) methods \cite{pugh2016quality}, try to maximize diversity in the discovered stepping stones that can lead to more and more complex artifacts \cite{lehman2011evolving,mouret2015illuminating}. Algorithms such as POET \cite{wang2019paired} extend these ideas to generate both the environments and agent behaviors, creating a system for more open-ended discovery. To see the full benefits  of EvoCraft, it will likely have to be combined with one of these QD algorithms in the future, or will more likely require the development of new open-ended learning methods. 

\subsection{Minecraft}
Minecraft has emerged as a popular testing environment for a diverse array of AI benchmarks. A popular framework for training reinforcement learning agents in Minecraft is Project Malmo \cite{johnson2016malmo}, which gives an abstraction layer on top of Minecraft that makes it straightforward to integrate AI agents and to design and run experiments. Project Malmo and connected research efforts such as MineRL \cite{guss2019minerl} (which includes a large dataset of human Minecraft demonstrations) focus on training agents operating from a first-person perspective to solve challenges such as vision, navigation, long-term planning, or interacting with other agents. In contrast, EvoCraft focuses on the idea of open-ended discovery, providing an API that allows the direct manipulation of blocks, instead of training agents to manipulate them. 


The closest related challenge to ours is the Minecraft AI Settlement Generation Challenge \cite{salge2018generative,salge2020ai}. In this challenge, the goal is to write an AI program that can create interesting settlements for unseen maps. In contrast to EvoCraft, the settlement API does not support reading information back from the environment. In other words, in the settlement generation challenge, Minecraft is used to render a settlement that is generated offline based on a given map description. Instead, EvoCraft  focuses on generating artifacts whose performance should be judged by the way they interact with the environment. 

Others have explored to train agents in Minecraft through interactive evolution to solve simple navigation tasks  \cite{de2018collaborative}, or created a Minecraft-like environment in which users can interactively evolve 3D building blocks and then collaboratively build larger structures \cite{patrascu2016artefacts}.

In summary, EvoCraft complements the existing Minecraft APIs that are used for research in AI, by enabling real-time feedback on the position and types of blocks, and by including redstone components that allow the evolution of mechanics and circuits. 

\section{The EvoCraft Environment}

EvoCraft is a mod for Minecraft that allows clients to manipulate blocks in a running Minecraft server programmatically through an API. A Python implementation of EvoCraft is available at: \href{https://github.com/real-itu/Evocraft-py}{github.com/real-itu/Evocraft-py}. To use the server interface with other languages, we provide interface definition files to generate clients for (almost) any programming language:  
\href{https://github.com/real-itu/minecraft-rpc}{github.com/real-itu/minecraft-rpc}. 

The API is based on the Sponge modding platform \cite{Sponge} that enables creating mods for Minecraft in form of Java plugins for the Minecraft server. Communication with the EvoCraft mod is based on gRPC Remote Procedure Calls (gRPC) framework which allows defining language agnostic strongly typed RPCs\footnote{https://grpc.io}. A gRPC specification consists of \texttt{services} and \texttt{messages}. A \texttt{service} exposes one or more \texttt{rpcs}, which are the callable endpoints and the \texttt{messages} define the inputs and outputs. A \texttt{message} is composed of other \texttt{messages} and primitive data types, e.g. strings, integers, etc. The EvoCraft   RPC specification completely defines the API and serves as the primary documentation.

The initial release of the EvoCraft API has a single \texttt{service} with three \texttt{rpcs}. See Listing~\ref{listing:service}. The Minecraft mod implements the EvoCraft gRPC API and listens on port 5001.

\begin{lstlisting}[language=protobuf2, style=protobuf, breaklines=false, caption={The EvoCraft gRPC API definition. For brevity the headers and message definitions are omitted.}, label=listing:service, captionpos=b]
service MinecraftService {
    /** Spawn multiple blocks. */
    rpc spawnBlocks (Blocks) returns (Empty);

    /** Return all blocks in a cube */
    rpc readCube (Cube) returns (Blocks);

    /** Fill a cube with a block type */
    rpc fillCube (FillCubeRequest) returns (Empty);
}
\end{lstlisting}

Given the gRPC definition, it is possible to automatically generate clients in multiple programming languages, e.g. Python, Java, Go, etc. The initial release contains a generated python client. See listing \ref{listing:python-example} for an example of using the python client.

An example of slime-block-based flying technologies, using a specific placement of pistons, blocks of redstone, observers, and slime blocks. A simple flying machine already requires a non-trivial arrangement of blocks. In Section~\ref{sec:flying_machine} we investigate how difficult it is for evolution to discover a moving machine from scratch.  A detailed description of the different block types is given in Appendix~\ref{appendix:block_types}.
\newpage

\begin{lstlisting}[language=python, style=protobuf, breaklines=false, caption={Example use of the EvoCraft python client to clear a $20\times10\times20$ working space (line 8) and spawn a minimal flying machine (line 15). See Figure~\ref{fig:flyer} for the resulting flying machine.}, label=listing:python-example, captionpos=b]
import grpc
import minecraft_pb2_grpc
from minecraft_pb2 import *

channel = grpc.insecure_channel('localhost:5001')
client = minecraft_pb2_grpc.MinecraftServiceStub(channel)

client.fillCube(FillCubeRequest(
    cube=Cube(
        min=Point(x=-10, y=4, z=-10),
        max=Point(x=10, y=14, z=10)
    ),
    type=AIR
))
client.spawnBlocks(Blocks(blocks=[
    # Lower layer
    Block(position=Point(x=1, y=5, z=1), type=PISTON, NORTH),
    Block(position=Point(x=1, y=5, z=0), type=SLIME, NORTH),
    Block(position=Point(x=1, y=5, z=-1), type=STICKY_PISTON, SOUTH),
    Block(position=Point(x=1, y=5, z=-2), type=PISTON, NORTH),
    Block(position=Point(x=1, y=5, z=-4), type=SLIME, NORTH),
    # Upper layer
    Block(position=Point(x=1, y=6, z=0), type=REDSTONE_BLOCK, NORTH),
    Block(position=Point(x=1, y=6, z=-4), type=REDSTONE_BLOCK, NORTH),
    # Activate
    Block(position=Point(x=1, y=6, z=-1), type=QUARTZ_BLOCK, NORTH),
]))
\end{lstlisting}


\begin{figure}
    \centering
    \includegraphics[width=2.9in]{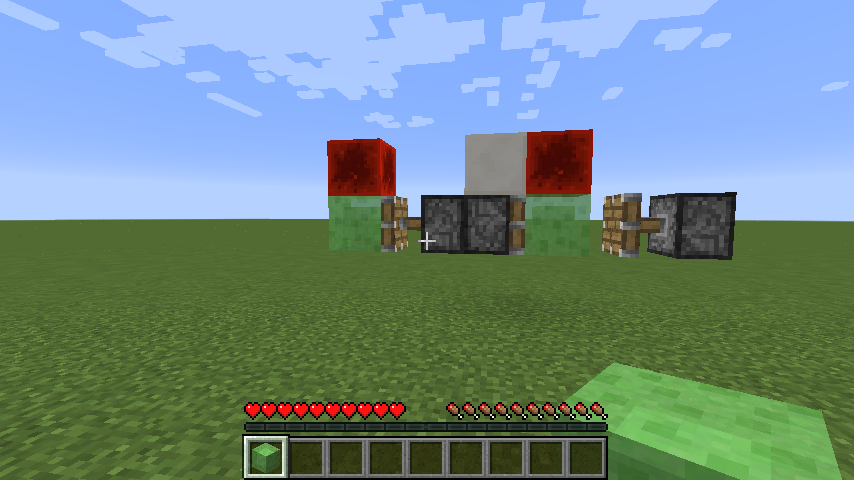}
    \caption{A minimal flying machine (created by code in listing~\ref{listing:python-example}), which perpetually moves north.}
    \label{fig:flyer}
\end{figure}

\subsection{Performance}
The Minecraft server runs at a fixed (maximum) update rate of 20 ticks/second. At each tick, the game state is updated and sent to the clients. If the server is overloaded such that a tick takes more than 50ms, the tick rate drops. We perform two tests to estimate how performant the EvoCraft is on a Macbook Pro with a 2.4 GHz 8 core Intel i9 processor and 16 GB 2400 MHz DDR4 RAM. Note that the Minecraft server only uses a single core since it is single-threaded.

\begin{figure}[ht]
    \centering
    \includegraphics[width=0.9\textwidth]{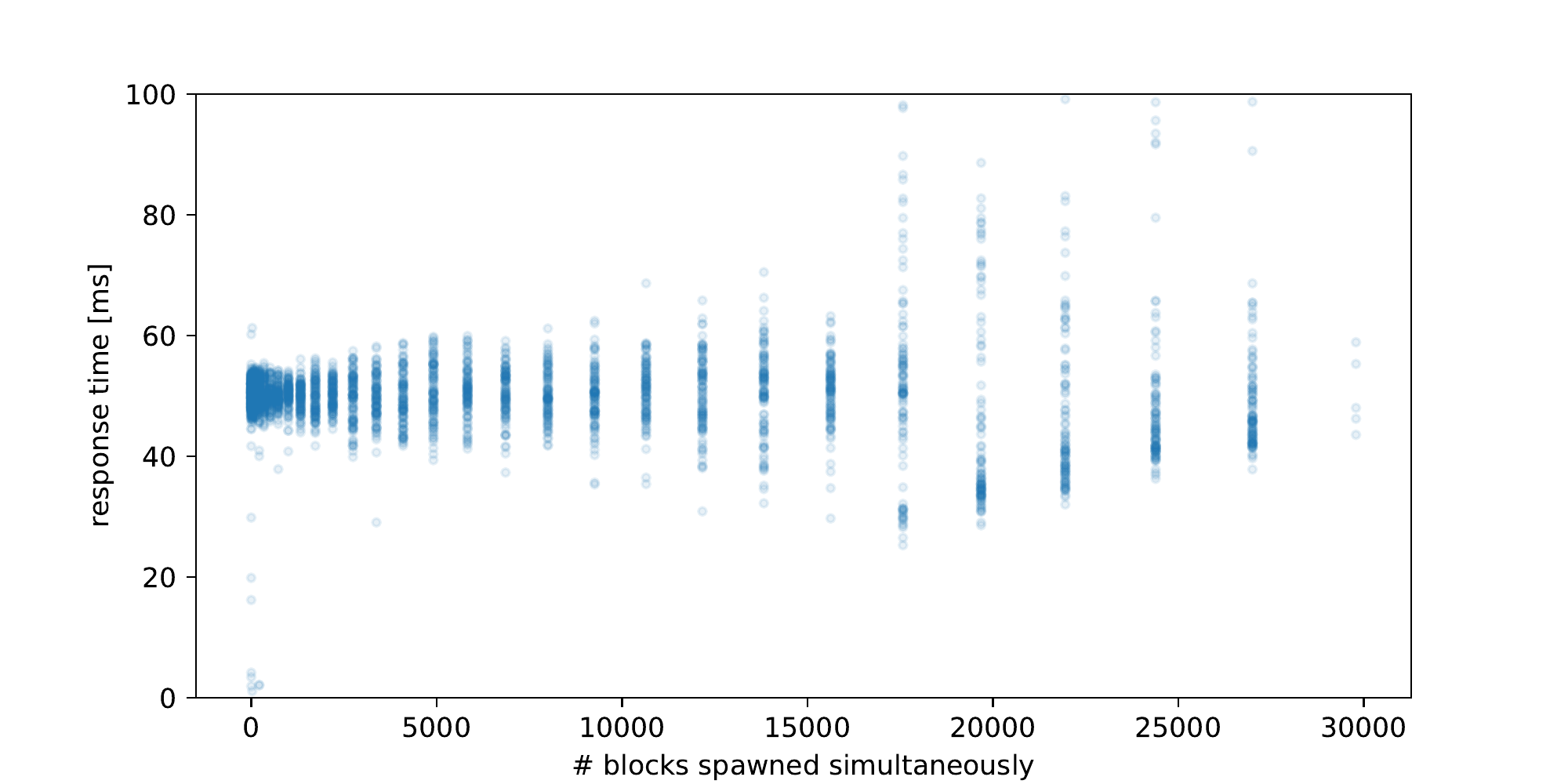}
    \caption{Response times when spawning blocks using the EvoCraft API.}
    \label{fig:cube-test}
\end{figure}

The first test repeatedly spawns an $N \times N \times N$ block of obsidian followed by an $N \times N \times N$ block of air for increasing values of $N$. When the server tick rate falls below 20 ticks/s the test is stopped. We measure the response time for the command to spawn the obsidian block, which is constant at approximately 50ms corresponding to a single tick until 29,791 blocks corresponding to a $31 \times 31 \times 31$ cube, at which point the tick rate drops and the test is stopped (Figure~\ref{fig:cube-test}).

In the second test, we simultaneously spawn $N$ minimal flying machines (Figure~\ref{fig:flyer})  and let the server run for 20 seconds and observe the tick rate. These machines are more demanding on the tick-rate than static blocks due to the dynamic behavior. We vary $N$ until the server tick rate falls below 20 ticks/s, which we observe happens when we spawn more than approximately 1200 flying machines.

Since Minecraft runs in a single thread it is often CPU-bound on modern multi-core hardware. To overcome this limitation, multiple instances of  EvoCraft can be run in parallel to scale the throughput of the simulation, although the client must balance the load manually among the environments.

\subsection{Relevant Minecraft Blocks}
In total, 254 blocks are available in this implementation.  However, many of them do not have a specific functional role. Below we focus on some essential block types, notably redstone circuits. Appendix~\ref{appendix:block_types} includes some  general comments about blocks in Minecraft, inlcuding information on basic blocks (Table~\ref{tab:blocks}), Redstone components (Tables \ref{tab:power}, \ref{tab:transmission}, \ref{tab:mechanism}), and a rough classification of the remaining blocks (Table~\ref{tab:transmission}).

\paragraph{Redstone Circuits} --which can be thought of as kind of electrical circuits-- enable to activate and control mechanisms and therefore to create machines (e.g.\ elevators, automatic farms, robots, CPUs, etc.), which can also be reactive to events in the environment (e.g.\  mob movement, item drops, etc.). Three types of elements are typically involved: 
\begin{enumerate}
\item[$(i)$] A power component provides power to all or part of a circuit, may it be transmission components or adjacent mechanisms components; e.g., Redstone torch, Button, Lever,  Detector rail, Pressure plate, Redstone block, etc. These items provide either short or continuous pulses, allowing mechanisms to be activated or de-activated (Table~\ref{tab:power}).
\item[$(ii)$] A transmission component passes power from one part of the circuit to another; e.g., redstone dust, redstone repeater, redstone comparator (Table~\ref{tab:transmission})
\item[$(iii)$] A mechanism component that once activated, changes  its state or something else in the environment (by producing light, moving, throwing something, etc.); e.g., Piston, Redstone lamp, Doors, Dispenser, etc. (Table~\ref{tab:mechanism}).
\end{enumerate}

\section{Evolutionary Optimization: Baseline Approaches}
Here we present some initial results on interactive (Section~\ref{sec:IEC}) and automated  evolution (Section~\ref{sec:automated_evolution})   with our API, showcasing some of its abilities. 

\subsection{Human in the loop: Interactive Evolution}
\label{sec:IEC}

In interactive evolutionary computation (IEC), a human may intervene to leverage the evolutionary processes, through various forms of guidance \cite{hutchison_interactive_2006,takagi_interactive_2001}, A major track within IEC is for the human to have some control (partial or full) over how fitness values are assigned. 
Consequently, IEC is particularly well suited  for domains or tasks for which it is hard to explicitly define a metric or, more crucially, when criteria are inherently subjective or ill-defined (e.g.\ beauty, salience, complexity, open-endedness). While it is hard to narrow down an analytical expression for these attributes, humans can be arguably apt at evaluating them. Interactive evolution seems notably a promising path to explore the quest for open-endedness \cite{secretan2011picbreeder,stanley2017open}.



We should mention that while the promise of IEC is appealing, there are undeniable drawbacks of this approach worth mentioning such as human-bias, fuzzy behaviors, or limited exploration of the search space. Furthermore, since a large number of evaluations may be required, it may trigger what is known as ‘user fatigue’, possibly leading to further noise and bias in the process.
Convergence behavior is, even with strong assumptions on the consistency of the user's behavior, rarely guaranteed, as it depends on the geometry of the desired phenotypic space. 

For these reasons, hybrid processes using partial human interventions (and possibly proactive instead of reactive) along with other methods may be preferred instead of full human guidance and prove to be more tractable. For example, interactive evolution has been combined with novelty search to increase searching efficiency \cite{lowe2016accelerating,woolley_novel_2014}. 

For the initial EvoCraft experiments in this paper, we restrict ourselves to the simplest implementation of interactive evolution. We use an evolutionary strategy \cite{Salimans2017Mar} to evolve entities and ask a human to choose its favorite among the candidate solutions at each generation.

\subsubsection{Artefact Encoding}
\label{art_enc}

The encoding is a  simple feedforward multi-layer-perceptron, composed of three linear layers of 20 hidden nodes each and non-linear activations. The network is queried on a bounded spatial zone (2D or 3D box) to produce a structure. 
The input is composed of the relative 3D spatial coordinates and the distance to the center, which can also be symmetrized beforehand. 
The output is either ($N$+1)-dimensional, or ($N$+7)-dimensional if the block orientations are included in the encoding, where N is the number of block types allowed.
While the first dimension determines if the queried position shall be air or matter, the following $N$-dimensions dictate the probability of using a certain type of block at this position. 
To encourage more symmetric creations, a variant of this encoding has been tested, where gaussian, tanh, sin and cos are replacing the default activations (relu and sigmoid). These activations, which resemble  CPPN-network activations \cite{stanley2007compositional}, enable more symmetric and smooth compositions.
However, due to the current discretization, and the simplicity of this first encoding, the evolved shapes do not have the organicity which famously characterizes CPPN-compositions \cite{secretan2011picbreeder,stanley2007compositional}.

While the current implementation allows for the generation of structures to be stochastic via a top-k-sampling, it is currently run deterministically  ($k=1$).

\subsubsection{IEC Setup}

We use an evolutionary strategy (ES)  \cite{Salimans2017Mar} to generate a set of candidates that are presented to a human judge who has to choose one among them. Fitness is set to 1.0 for the chosen entity and 0.0 for all others. Subsequently, a new generation of candidate solutions is produced by the ES algorithm based on these fitness values. This process can be repeated ad infinitum. 

The ES algorithm uses a noise modulating parameter $\sigma \sim 0.1$ and a small learning rate $\sim 0.01$ in order to obtain a gradual evolution so that users can make sense  of the phenotypic changes produced by their choices. To speed up the evolutionary process, structures generated below a minimal size of blocks are not displayed, and their fitness is automatically set to 0.


\subsubsection{Results}
\begin{figure}[h]
     \centering
     \begin{subfigure}[b]{0.3\textwidth}
         \centering
         \includegraphics[width=\textwidth]{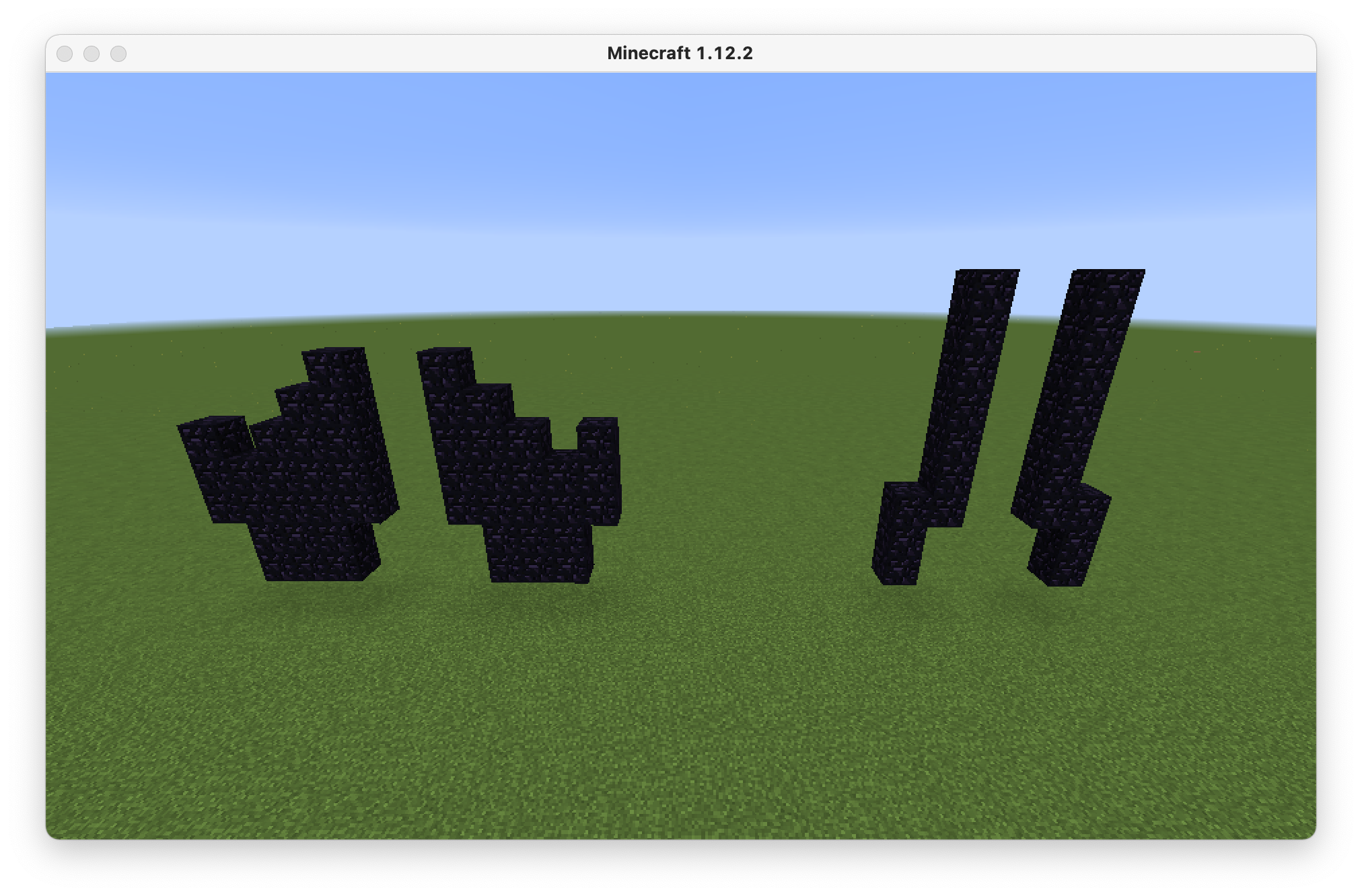}
     \end{subfigure}
     \begin{subfigure}[b]{0.3\textwidth}
         \centering
         \includegraphics[width=\textwidth]{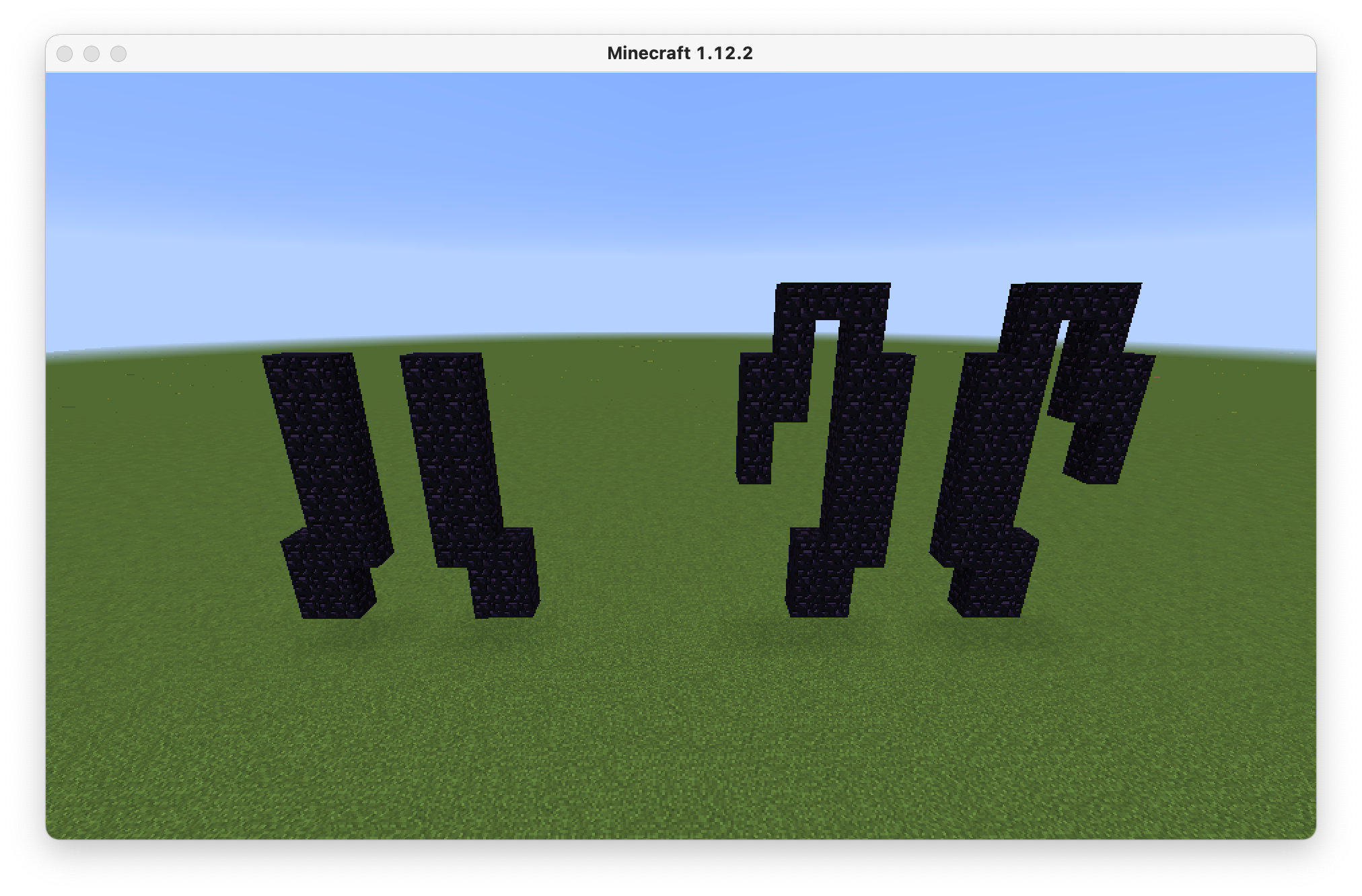}
     \end{subfigure}
     \begin{subfigure}[b]{0.3\textwidth}
         \centering
         \includegraphics[width=\textwidth]{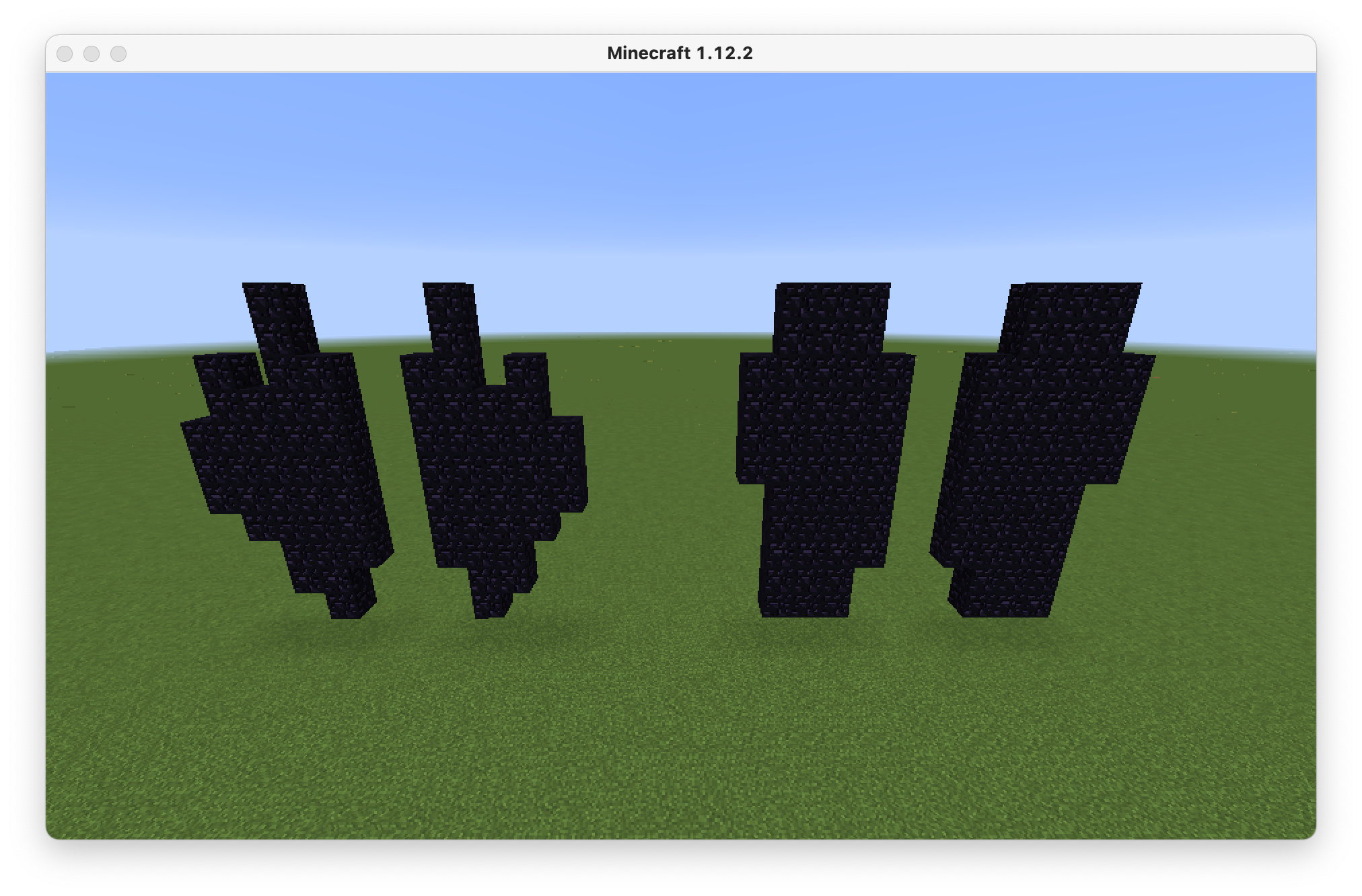}
     \end{subfigure}
     \centering
     \begin{subfigure}[b]{0.3\textwidth}
         \centering
         \includegraphics[width=\textwidth]{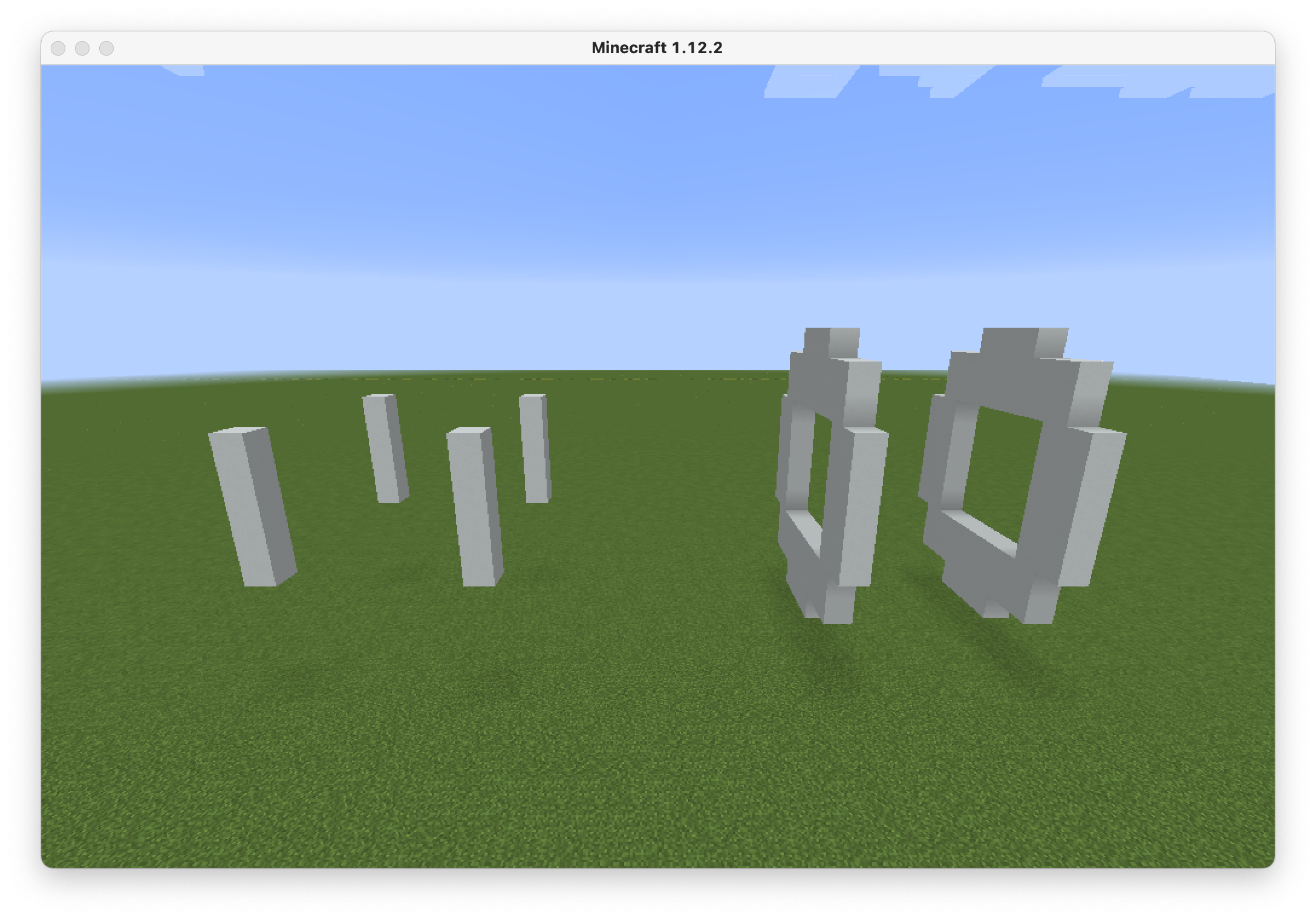}
     \end{subfigure}
     \begin{subfigure}[b]{0.3\textwidth}
         \centering
         \includegraphics[width=\textwidth]{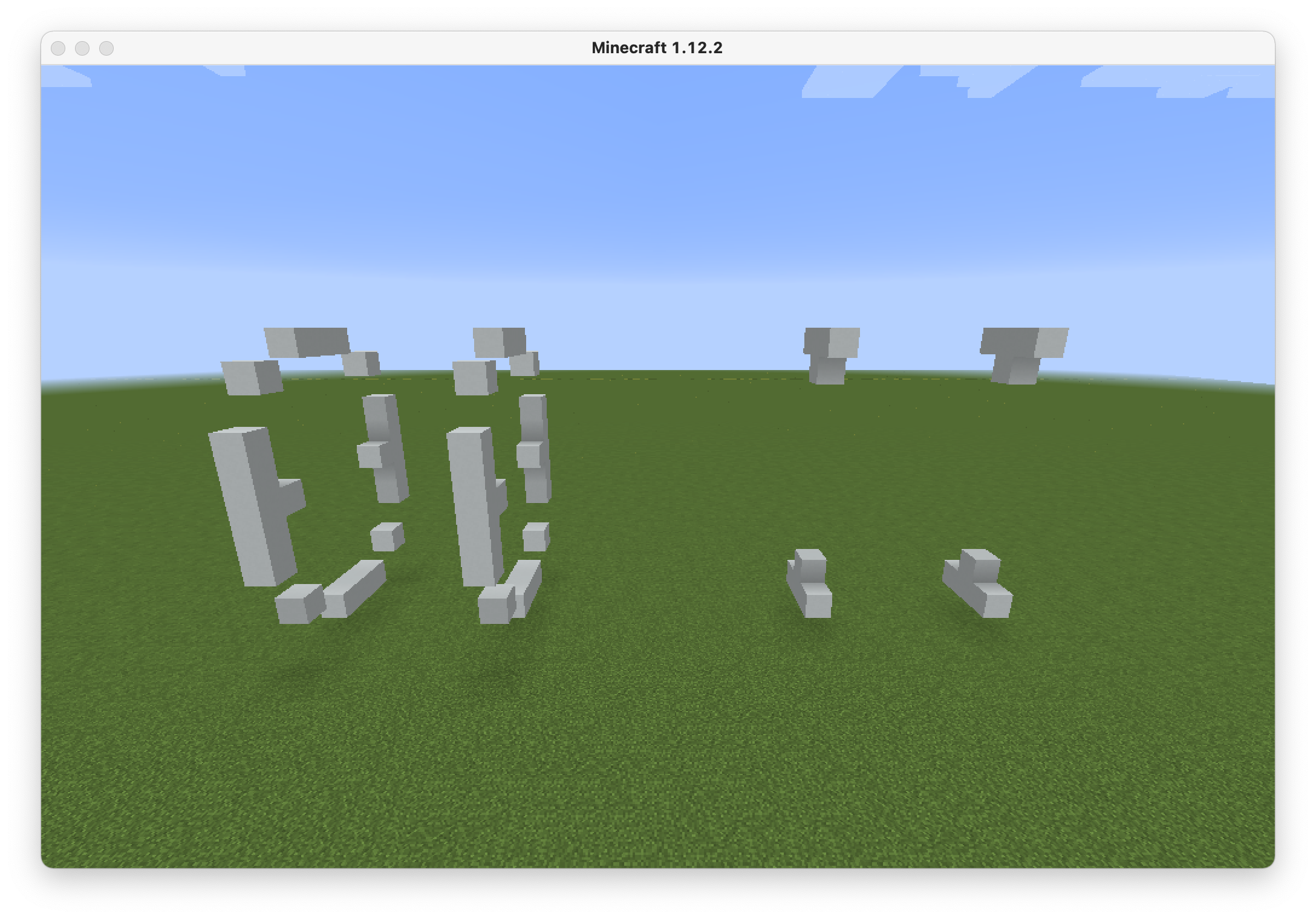}
     \end{subfigure}
     \begin{subfigure}[b]{0.3\textwidth}
         \centering
         \includegraphics[width=\textwidth]{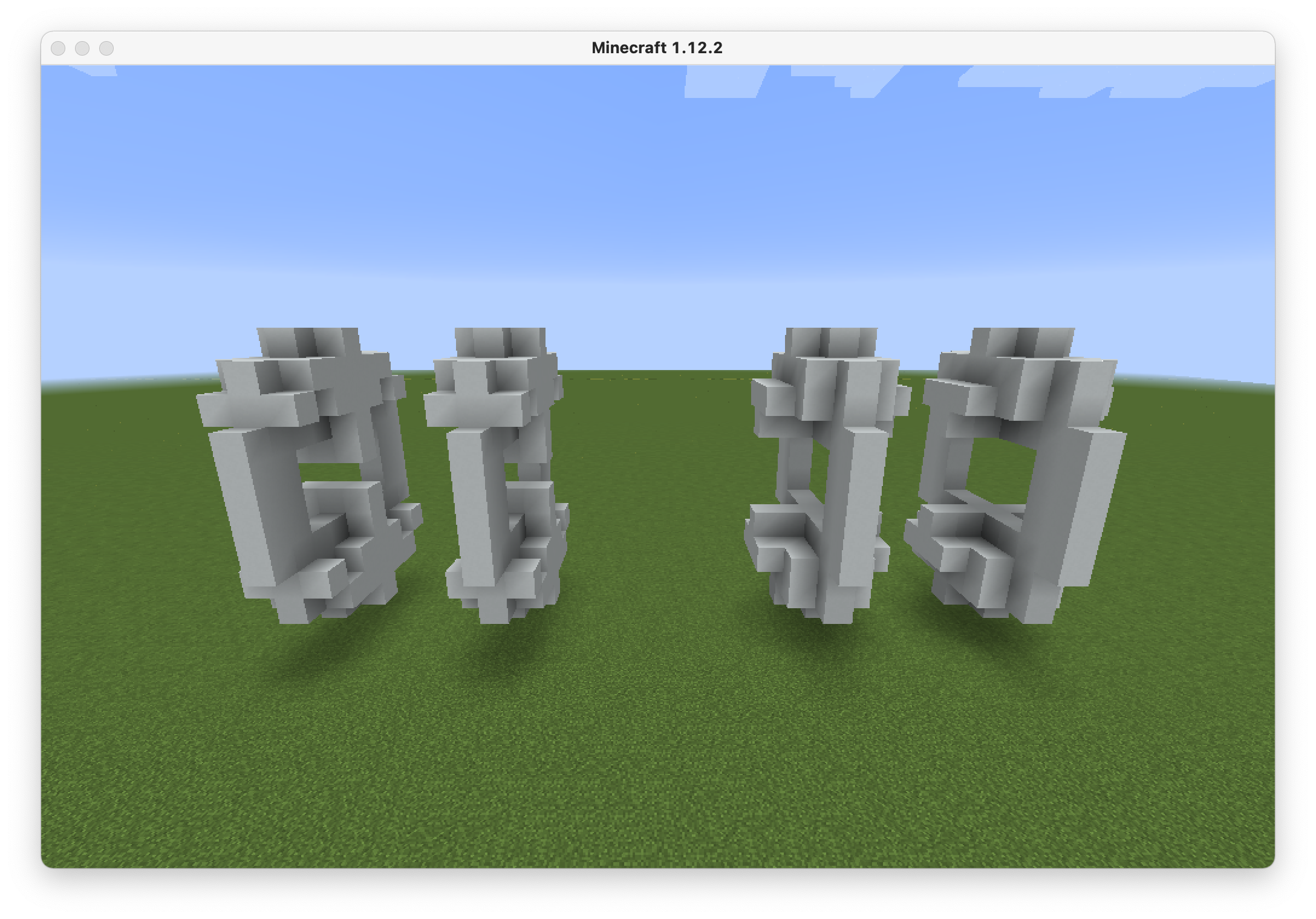}
     \end{subfigure}
     \centering
     \begin{subfigure}[b]{0.3\textwidth}
         \centering
         \includegraphics[width=\textwidth]{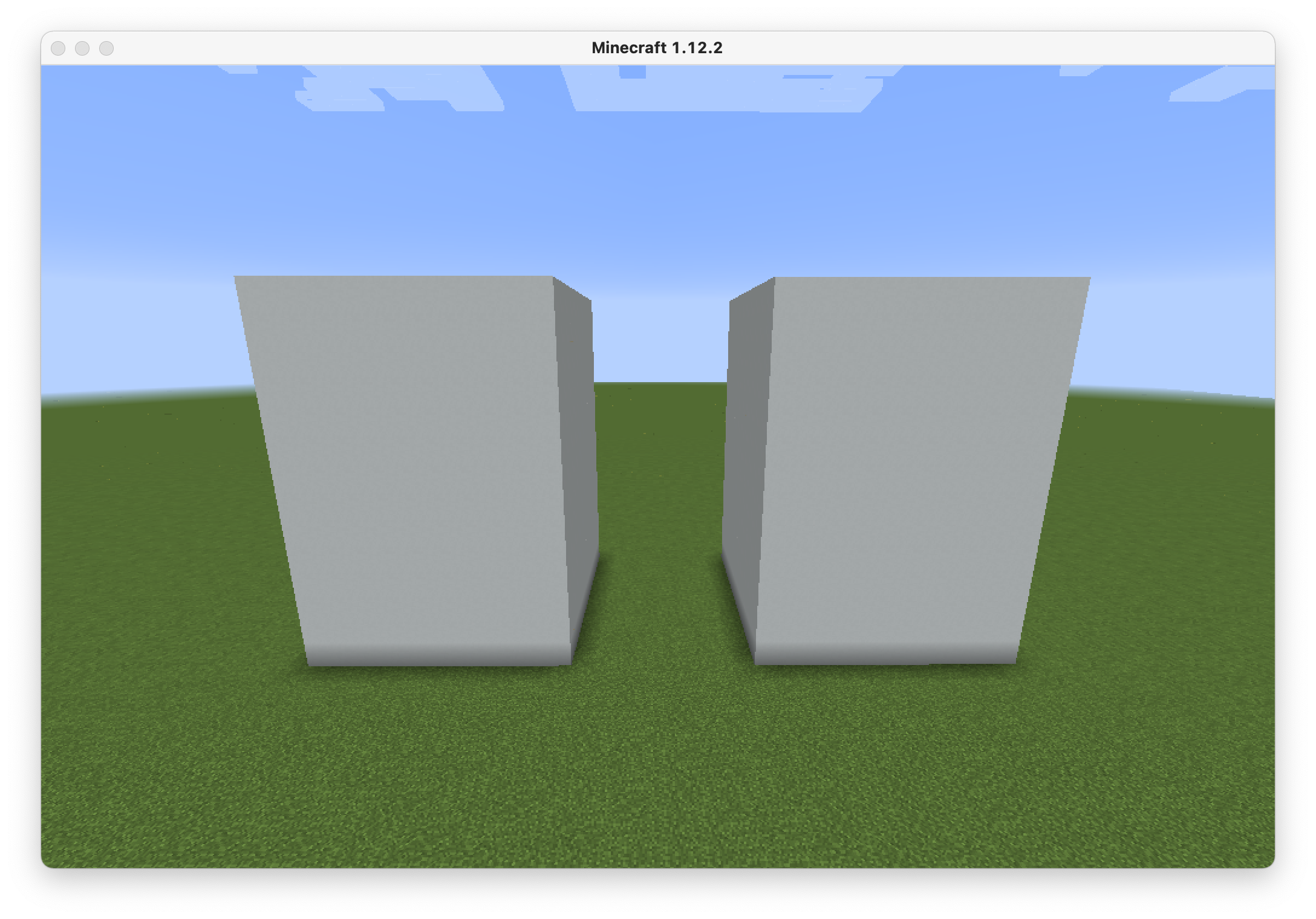}
     \end{subfigure}
     \begin{subfigure}[b]{0.3\textwidth}
         \centering
         \includegraphics[width=\textwidth]{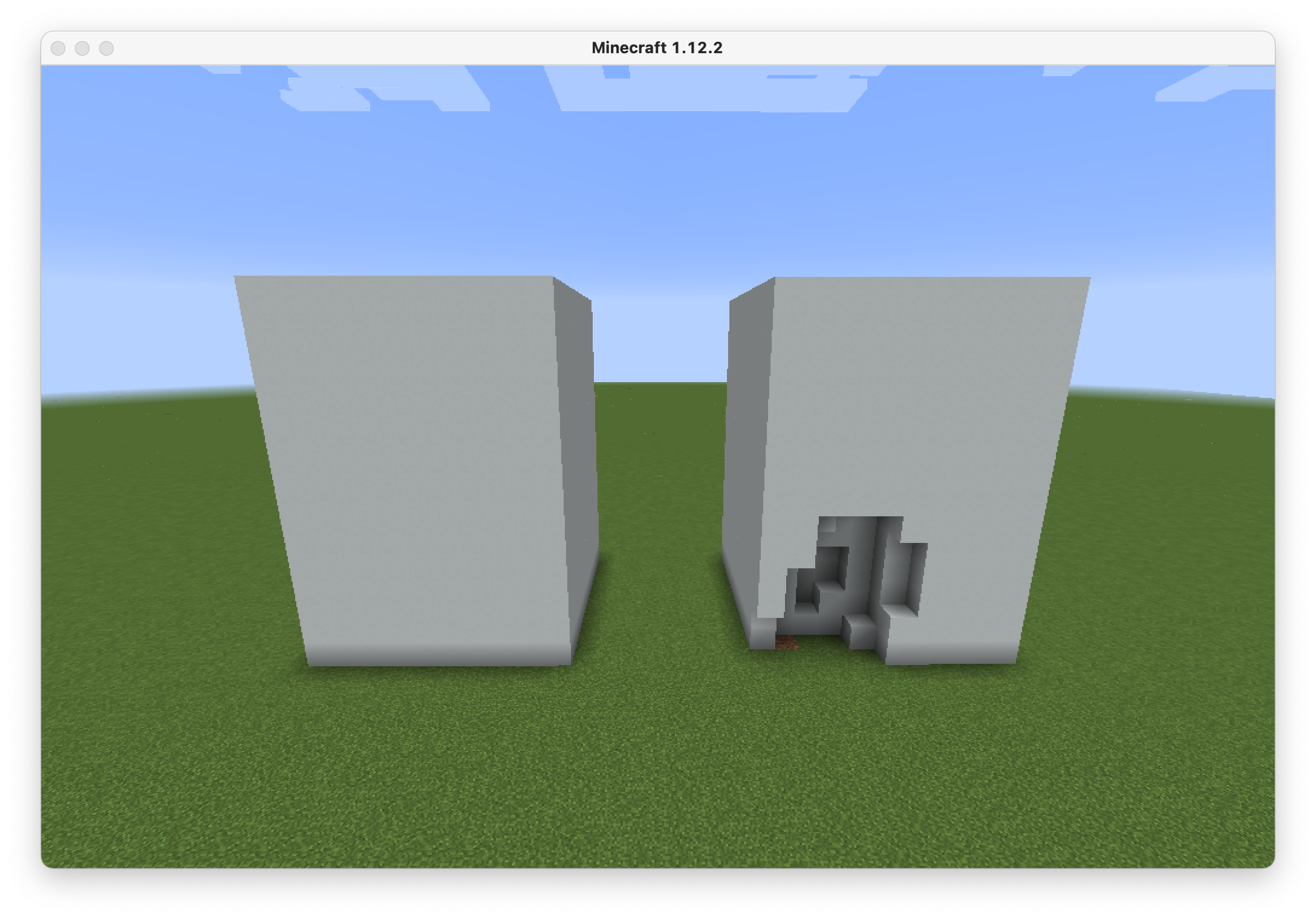}
     \end{subfigure}
     \begin{subfigure}[b]{0.3\textwidth}
         \centering
         \includegraphics[width=\textwidth]{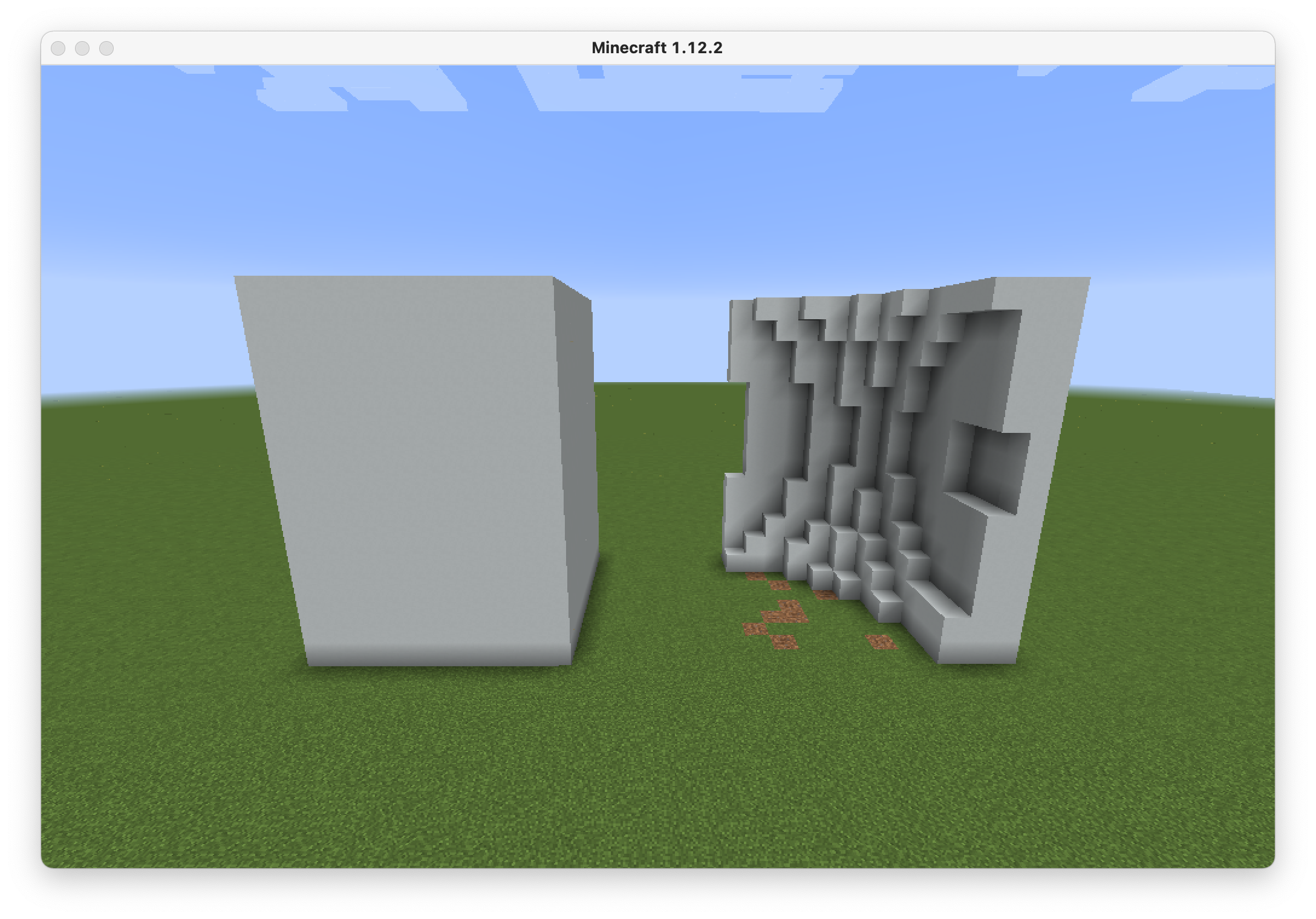}
     \end{subfigure}
     \centering
     \begin{subfigure}[b]{0.3\textwidth}
         \centering
         \includegraphics[width=\textwidth]{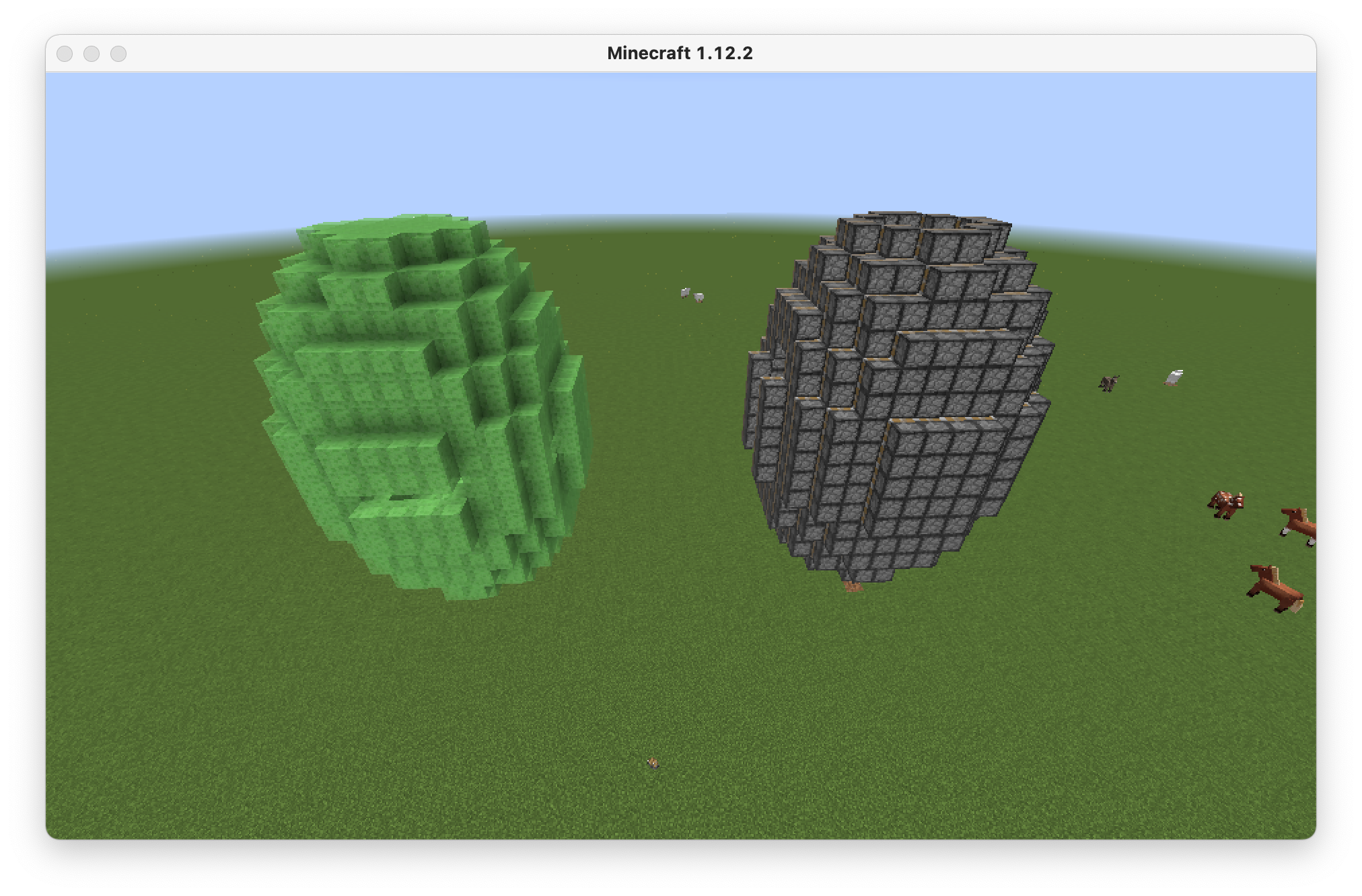}
     \end{subfigure}
     \begin{subfigure}[b]{0.3\textwidth}
         \centering
         \includegraphics[width=\textwidth]{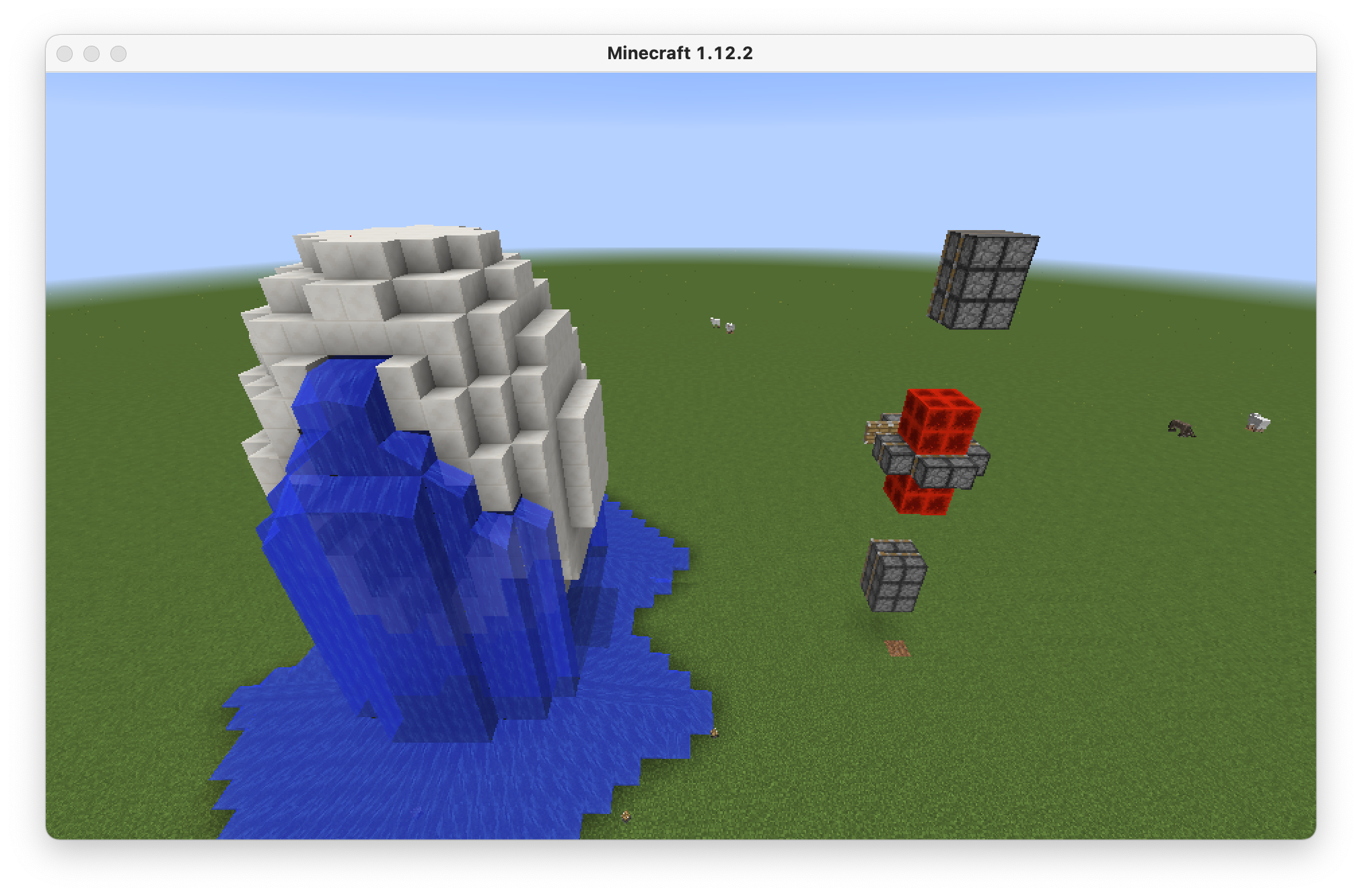}
     \end{subfigure}
     \begin{subfigure}[b]{0.3\textwidth}
         \centering
         \includegraphics[width=\textwidth]{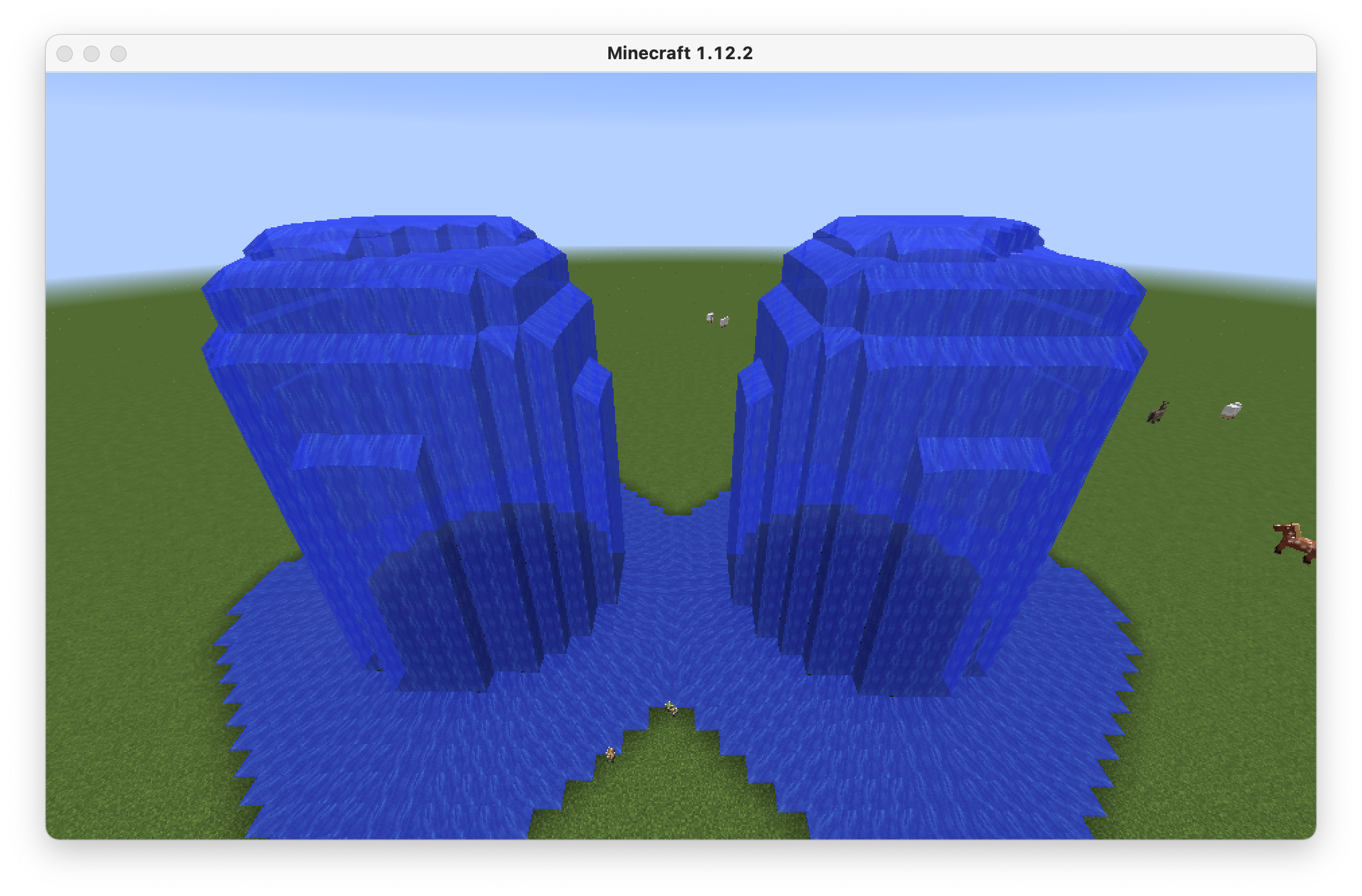}
     \end{subfigure}
    \hspace{\bibindent}\raisebox{0cm}{Generation 0 \hspace{1.5cm} Generation 10 \hspace{1.5cm} Generation 30}
     \caption{Each row shows a different evolutionary run where the evolution of the artifacts is guided by the choice of a  human experimenter. In the first three experiments, the human was asked to choose the shape they found most interesting while in the last one they were asked to produce a waterfall-like entity. Each experiment was run over 30 generations, the first being the left column and the last the right column.}
          \label{fig:human_evolved_artefacts}
\end{figure}

The entities evolved using IEC are shown in Figure~\ref{fig:human_evolved_artefacts}. Each row shows an evolutionary run at three stages; the left column displays the entities from the  first generation and the right column  the last generation. In the first three rows the user (one of the authors) was asked to choose whichever entities they found more interesting. On the other hand, for the last's row experiment, the goal was ``to build a waterfall''. Here the user selected over generations for entities which had water blocks. Eventually, the solutions converged to generating waterfall-like entities.

Let us underline the importance of the chosen genotype-to-phenotype encoding on the artifacts evolved, which may hinder the evolutionary road towards certain traits or artifacts while favoring others; representation learning is as crucial for open-endedness as it is for standard machine learning. 
The provided evolved examples (Figure~\ref{fig:human_evolved_artefacts}) serve as evidence that interactive evolution can be used to evolve interesting artifacts in a controlled manner in EvoCraft. However, the resulting evolved objects tell more about the underlying encoding representation than about the process of interactive evolution itself.


\subsection{Automated Evolution}
\label{sec:automated_evolution}
This sections presents two simple automated evolutionary experiments, which showcase how information collected from the game (e.g.\ the position of a particular block type) can be used to compute relevant fitness metrics.

\subsubsection{Evolving a tower growing towards a block of gold} In this experiment we evolve a tower structure that has to reach a golden block suspended in the sky. The tower is constructed out of a subset of all possible blocks, namely: obsidian, redstone, glass, brown mushroom, nether block, cobblestone, and slime blocks. The air blocks are not considered as a part of the tower.

We maintain a population of twenty towers and their respective targets, all existing at the same time in the same Minecraft world and being evaluated in parallel. Information on the position of the target is provided by the EvoCraft API. The distance between the closest tower block and the gold block is the tower's fitness value. The tower's initial block spawns at the base point. Each towers' target gold block is placed 10 block distances in the north direction, 10 distances in the west direction, and 10 distances in the up direction from the base point. 
For the next generation, we select the best 10\% of the towers as parents. Two random parents are selected and merged into a child tower. There is a 5\% chance that a child tower will mutate before it joins the next generation. The best tower in the current generation is copied over to the  next generation unchanged.

Instead of the neural network-based encoding used above, towers are encoded using a simple ternary tree, where each node represents a block (block type and facing direction) and three edges toward children nodes represent the direction (north, west, and up) in which the child node is connected to the parent node.  
Initial trees are randomly generated by taking a root node and generating a child in each cardinal direction with the probability of 0.5. Each next layer is generated recursively in the same way where the probability of creating children drops by 0.05 for each level. During the crossover of two parents, both parent trees are cut at a random child node; the first parent loses the sub-tree from the random cut toward the leaf, which is replaced by a sub-tree from the second parent.

\begin{figure}[h]
\captionsetup[subfigure]{labelformat=empty}
     \centering
     \begin{subfigure}{0.25\textwidth}
         \centering
         \includegraphics[width=\textwidth]{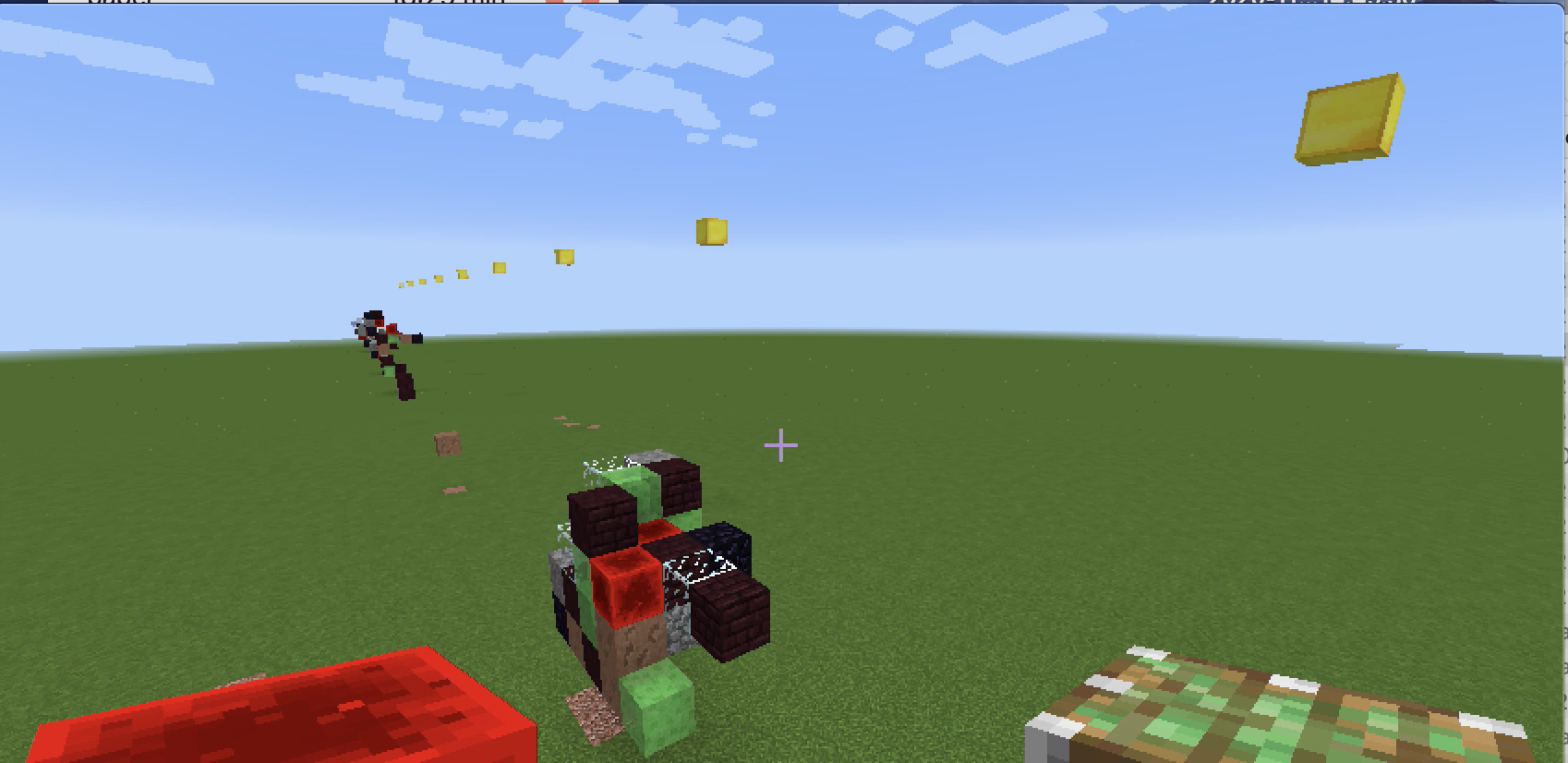}
         \caption{Generation 1}
     \end{subfigure}
     \begin{subfigure}{0.25\textwidth}
         \centering
         \includegraphics[width=\textwidth]{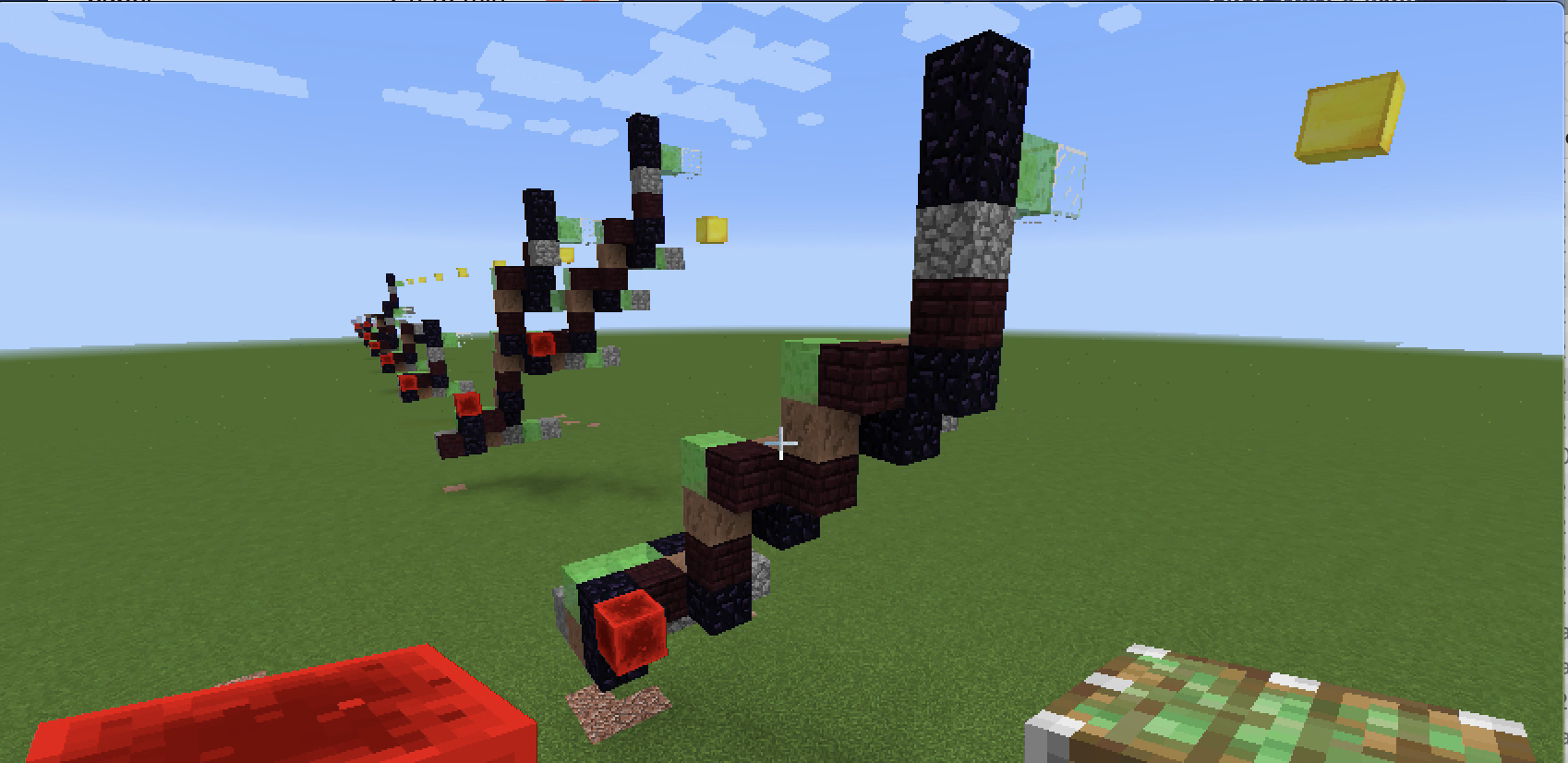}
         \caption{Generation 6}
     \end{subfigure}
     \begin{subfigure}{0.25\textwidth}
         \centering
         \includegraphics[width=\textwidth]{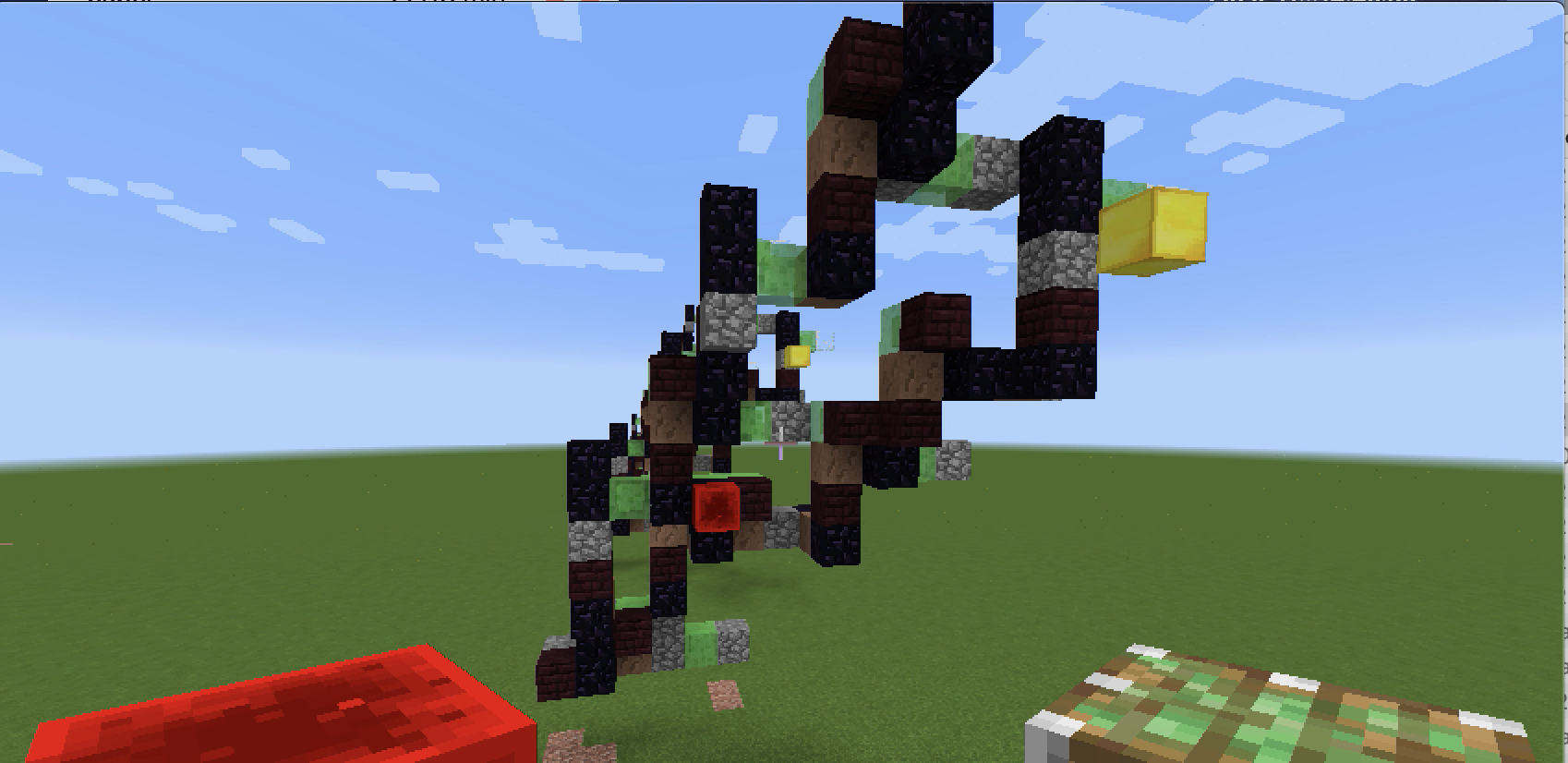}
         \caption{Generation 13}
     \end{subfigure}
          \begin{subfigure}{0.22\textwidth}
         \centering
         \includegraphics[width=\textwidth]{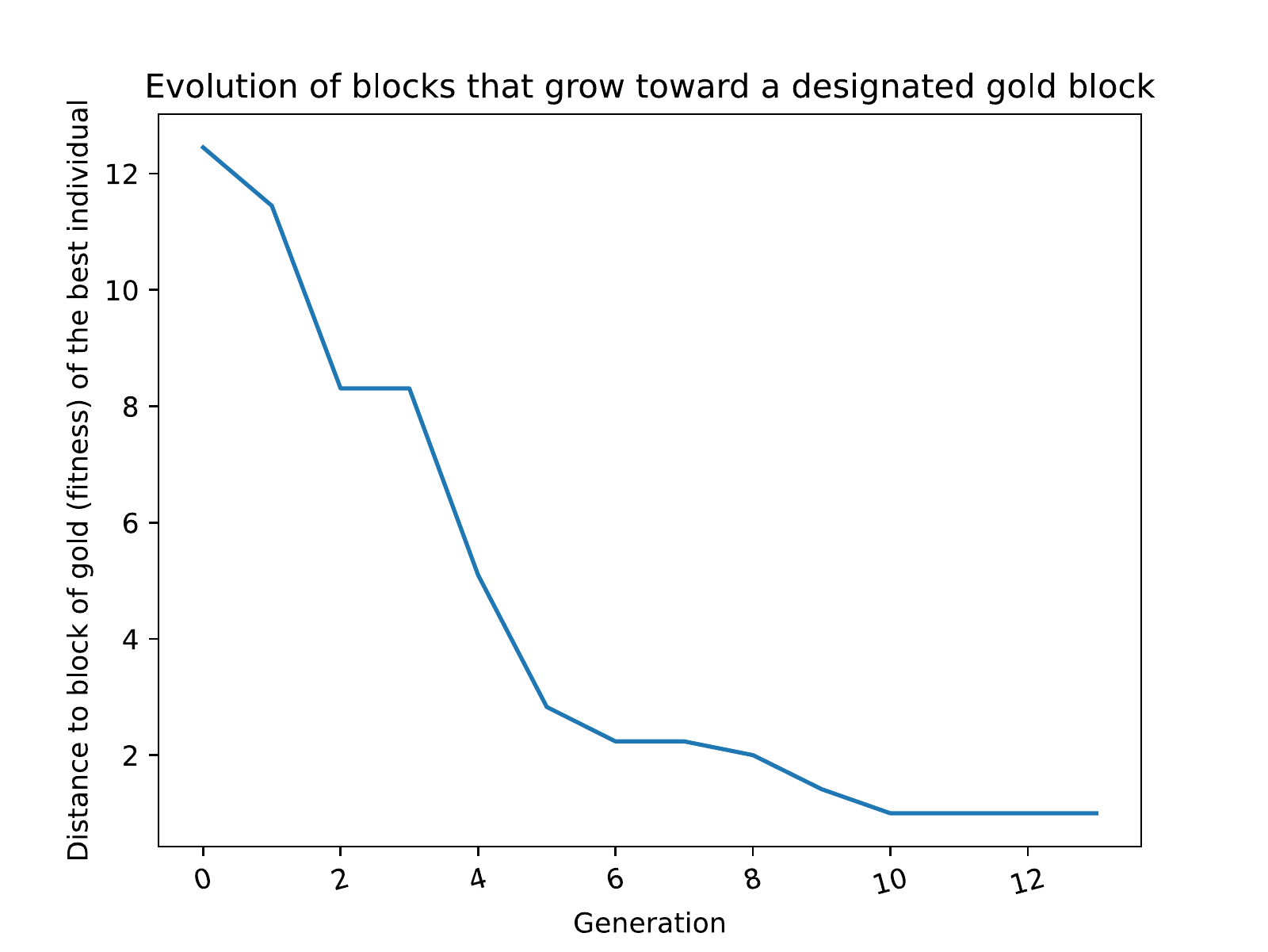}
         \caption{Fitness}
     \end{subfigure}
     \caption{Towers evolving to reach the golden block at generations 1, 6, and 13. Training performance over generations (minimum distance to the gold block) is shown on the right.}
     \label{fig:evolving_towers}
\end{figure}

Images of the best towers from the 1st, 6th, and 13th generation are shown in Figure~\ref{fig:evolving_towers}. Evolution is quickly able to find a tower design that reaches the target. Listing~\ref{listing:python-evol} in the appendix contains a minimal Python example on how to use the EvoCraft API to run this simple evolutionary loop.

\subsubsection{Evolving a moving machine}
\label{sec:flying_machine}

To demonstrate how the API can be used to track and evolve redstone machines, we attempted to evolve a moving machine, similar to the flying one shown in Figure~\ref{fig:flyer}. 
For this experiment, we again use the simple neural network  encoding (Section~\ref{art_enc}) together with an evolutionary strategy optimizer and a population size of ten. Fitness is calculated as the change in center of mass of the evolved structure after ten seconds. 

\begin{figure}[h]
     \centering
     \begin{subfigure}[b]{0.45\textwidth}
         \centering
         \includegraphics[width=\textwidth]{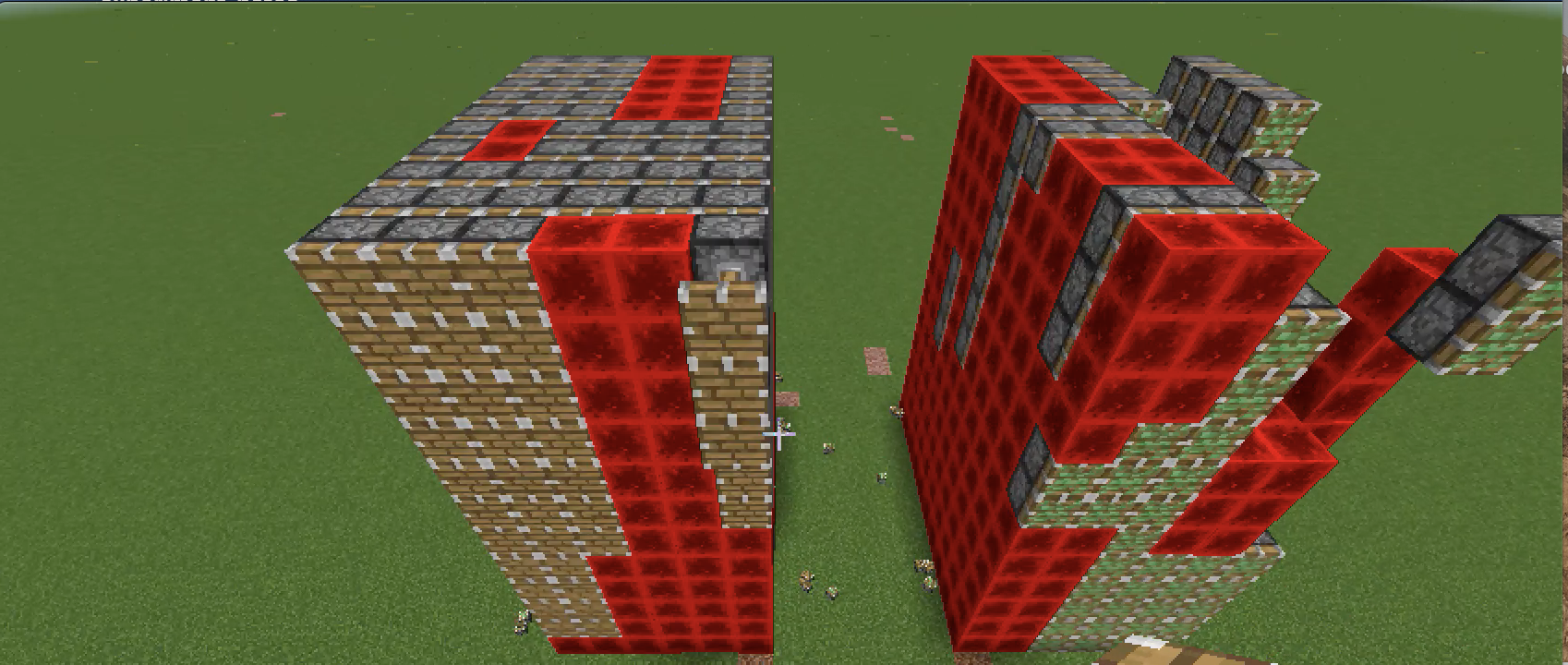}
     \end{subfigure}
     \begin{subfigure}[b]{0.45\textwidth}
         \centering
         \includegraphics[width=\textwidth]{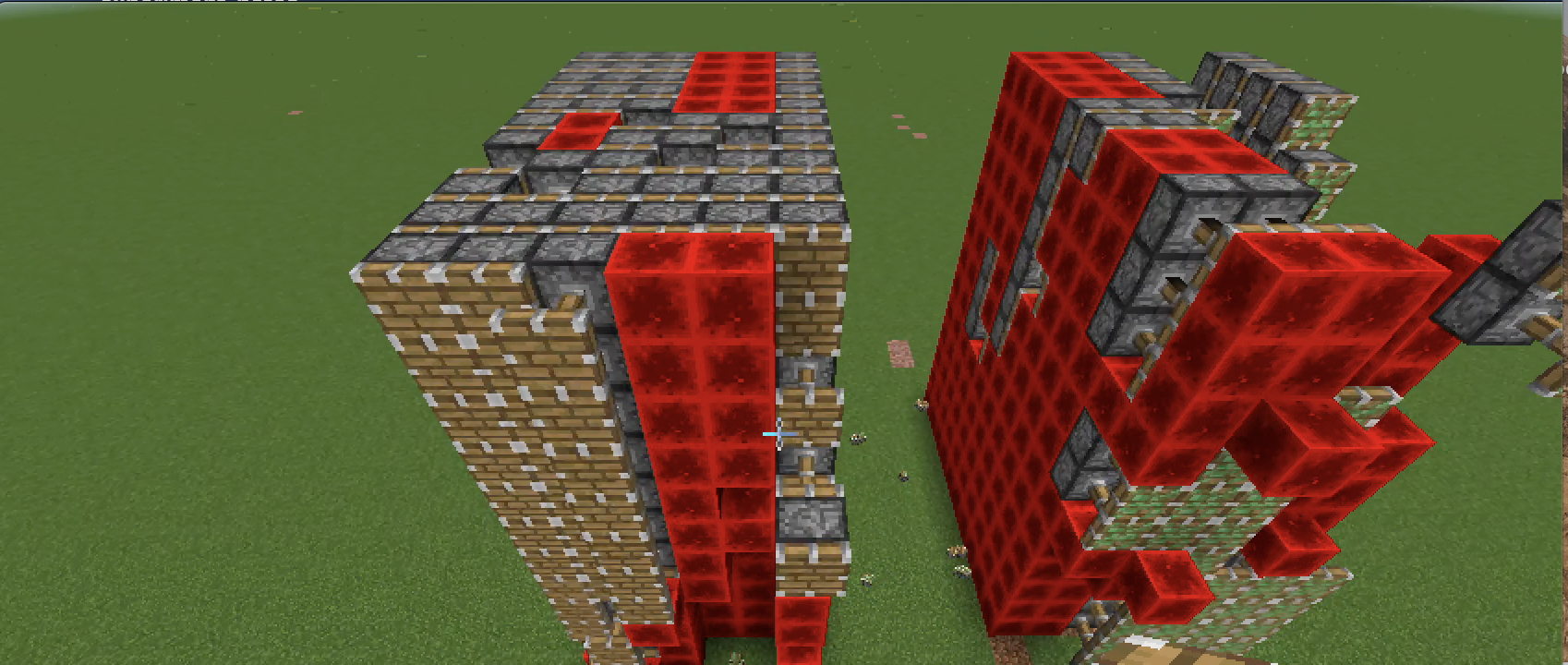}
     \end{subfigure}
     \caption{\textbf{Machines evolved for moving.} Two examples on the left  show machines before the triggered pistons push a part of the structure. The right figure shows the same machines after the triggered pistons displace some of the blocks. Evolution failed to find a machine that can perpetually move.}
     \label{fig:moving_machines_examples}
\end{figure}

Evolution was able to find some machines that do move their center of mass slightly. However, 
the presented machines are not consistently moving in one direction, as the flying machine does, but spawn a combination of redstone blocks adjacent to piston blocks that push parts of the structure to a different position (Figure~\ref{fig:moving_machines_examples}). The minimal flying machine displaces its center of mass by 17 positions in ten seconds. So far, we were able to find machines that can push their center of mass by 0.2 block positions after which they stop, meaning we are a long way from evolving a functional flying machine.



\section{Conclusion and Future Directions}
In this paper, we presented the first version of EvoCraft, a framework that allows the evolution of artifacts in Minecraft, including circuits and mechanics. Importantly and in contrast to the existing Minecraft AI frameworks, the provided API allows to easily place and track blocks in a running Minecraft server programmatically. Compared to related challenges, such as the Minecraft Settlement Generation Challenge, EvoCraft is more about -- but not exclusively focused on -- the evolution of mechanical/functional artifacts. We demonstrate the usability of EvoCraft on automated and interactive evolutionary tasks. 

Given the failure of purely fitness-based evolution in EvoCraft to evolve still relatively simple flying machines, a natural next step would be to try quality diversity algorithms \cite{pugh2016quality}. An interesting question is if these algorithms would be able to invent stepping stones that ultimately lead to interesting artifacts such as flying and self-replicating machines, CPUs, and word processors in Minecraft (Figure~\ref{fig:minecraft_creations})? Such experiments will help us to determine which ingredients are missing from current QD methods to create a system that can produce increasingly complex artifacts in complex worlds. 

While the initial experiments presented in this paper employ evolutionary optimization, EvoCraft is not restricted to such search methods. It will be interesting to see how other machine learning methods that could help in achieving open-endedness perform and compare to evolutionary approaches, such as self-play \cite{silver2017mastering} or methods for procedural content generation through reinforcement learning \cite{khalifa2020pcgrl}. Our framework should be able to naturally support these methods.

In addition to more powerful and exploratory search methods, another important next direction is to explore and develop different genetic encodings for EvoCraft. In this paper, we showed results with two different but relatively simple encodings. The field of generative and developmental systems \cite{stanley2003taxonomy} has produced many encodings such as neural cellular automata \cite{mordvintsev2020growing} or CPPNs \cite{stanley2007compositional} that could extend the diversity of artifacts we can evolve. Additionally, we imagine that encodings that can build a library of interesting building blocks (e.g.\ a moving part, a CPU register, etc.) will be essential to allow truly open-ended innovation in Minecraft.  


Important future work also includes extending the API in the next generation of the framework. For example, while we can simulate multiple individuals in the same Minecraft world, in the future we plan to add the ability to also parallelize evolution across multiple different Minecraft servers.


\section*{Acknowledgments}
We thank Christoph Salge, Raluca D. Gaina, and Sam Devlin for helpful discussions on Minecraft.  This project was partially supported by a Sapere Aude: DFF-Starting Grant (9063-00046B) and by the Danish Ministry of Education and Science, Digital Pilot Hub and Skylab Digital.
\bibliographystyle{splncs04}

\bibliography{refs}

\newpage
\appendix 

\section{Appendix: Minecraft Blocks}\label{appendix:block_types}
\subsection{General Blocks}
\label{sec:block_types}
Before providing an engine-oriented selection of block types  (cf. Table \ref{tab:blocks},\ref{tab:other_blocks}, \ref{tab:power}, \ref{tab:transmission}, \ref{tab:mechanism}), let us state a few general comments about Minecraft blocks:
\begin{itemize}
\item Almost all blocks ignore gravity, apart from a few blocks such as sand, gravel, anvil, snow, concrete powder.
\item Certain blocks can not be moved by players or entities, such as fire, lava, water, web.
\item Certain blocks require a particular tool (e.g. pickaxe or shovel) in order to be broken, e.g. Iron, Anvil, Rock, Snow, Brick. 
\item Some blocks may exist under different states --'Event' blocks-- e.g. (Flowing) lava, (Lit) Redstone Lamp, Powered/Unpowered Comparator, etc.
\item There are only two liquids blocks (water and lava), and they may obstruct movement. If combined, they turn into stone (when lava flows into water) or obsidian (if water flows into a lava source).
\end{itemize}

\begin{center}
\begin{table}
\caption {Minecraft Basic Blocks} \label{tab:blocks} 
\footnotesize
\begin{tabular}{ |p{2cm}|p{11cm}| } 
\hline
Air & Unbreakable and transparent block, air works as a substitute for the absence of blocks, allowing the player to walk through the world.\\
\hline
Fire &  Fire spreads to any close enough flammable block (e.g. Cloth, Vine, Wood, Leaves, etc.) and is extinguished by Rain. \\
\hline
Water (/Flowing) & Unlike lava, blocking or destroying a water source block stops the flow of water and removes the water that came from it. Pouring water on lava turn flowing lava into cobblestone, and lava source blocks into obsidian.\\
\hline
Lava (/Flowing) & Lava destroys any dropped blocks coming into contact with it. Blocking or destroying a lava source block does not stop the whole lava flow, which may set on fire things on its way. Getting rid of lava requires placing and destroying many blocks on top of the lava. \\
\hline
Stone (/Cobblestone) & Stone makes up most of the Minecraft world. Mining stone turns it into Cobblestone, which can, in turn, give back stone if burnt in a furnace.\\
\hline
Obsidian &  Obsidian may be created when water enters in contact with lava source blocks and is one of the toughest available elements, resistant to all explosions. \footnotemark[1]\\
\hline
Glass & Glass are transparent blocks that can be created by smelting sand in a furnace.\\
\hline
Concrete (/Powder) & Concrete is formed when concrete powder comes into contact with a block of water.  \\
\hline
Slime & Slime may designate bouncy cube-shaped hostile mobs, but is also an essential material in moving machine. It can form slime-based flying machines (as in \ref{fig:flyer}) when coupled with pistons and redstone blocks notably.\\
\hline
Bedrock & Bedrock --playing the role of world's boundaries-- is an indestructible block which covers the bottom of the Overworld and the top and bottom of the Nether.\\
\hline
\end{tabular}
\end{table}
\end{center}
\footnotetext[1]{Obsidian is used to create Nether portals, used to travel to the Nether, and enchantment tables, which opens to another side of Minecraft.}

\begin{center}
\begin{table}
\caption {Other Minecraft Blocks} \label{tab:other_blocks} 
\footnotesize
\begin{tabular}{ |p{3cm}|p{11cm}| }
\hline
Minerals, Metals &  They exist in form of Block and Ore which can be mined to produce resources \footnotemark[1]; e.g. Emerald, Diamond, Gold, Quartz, Coal, Gold, Lapis, Iron, Redstone.\\
\hline
Construction & Different types of Doors, Rails, Boxes, Chests, Stairs, Barriers, Fence, and Gates are available.\\
\hline
Slabs & Slabs are blocks that are half a block high. A few variants exists based on Stone, Purpur, or Wood possibly doubled.\\
\hline
Decorative & Some decorative objects, such as Bed, Bookshelf, Cauldron, Carpet, Cake, Flower Pot, Jukebox, Banner (Wall/Standing), Sign (Wall/Standing), Skull, Iron Bars, Ladder, Brewing Stand, Sea Lanter, Torch.\\
\hline
Cultivable & Certain things may be farmed in Minecraft, such as Wheat\footnotemark[3], Carrots,  Beetroots, Potatoes, Melon or Pumpkin, Sugar Cane (called 'REEDS' \footnotemark[4]), Cactus, Mushrooms, Cocoa, Trees (from 'Sapling'), Tallgrass, Flowers, Vines, etc.\\
\hline
Biosphere & Some other elements of biosphere present in Minecraft: Farmland \footnotemark[5], Dirt, some Plants (Chorus Plant, Double plant, Deadbush, etc), Leaves (and leaves2), Grass Path, Waterlily, Web, Sponge, some Egg, Mycelium, Wool, Bone block\footnotemark[6], etc.\\
\hline
Non-essential & Different forms of Terracotta, Glass (Pane/Stained/Stained Glass pane), Ice (possibly Packed/Frosted), (/Hardened) Clay (possibly Stained)\footnotemark[2], Log (and log2), Gravel, Planks, Purpur (Block/Pillar), Glowstone, Snow (and Snow Layer), Prismarine, Sand (/Soul, Red/Sandstone),  Magma\footnotemark[7].\\
\hline
Special Items  & A furnace (possibly Lit) allows players to cook their food and turn their ores into ingots; a torch is a light source which also prevents hostile mobs from spawning (as a Mob Spawner); a crafting table is used to craft items requiring more than the 4 crafting spots one have in the inventory display; an enchanting table may enchant weapons, tools, and armor; Anvil is used to repair items and combine enchantment; Structure Block may be used to manually generate, save or load structure, along with side Structure Void Blocks; Nether Blocks (Nether Brick, Fence, Stairs, Wart, etc.) are belonging to Nether World, End Blocks (End Portal, Bricks, Gateway, Stone, etc.) may be found in the end dimension\\
\hline
\end{tabular}
\end{table}
\end{center}
\footnotetext[1]{Apart iron and gold ores which have to be smelted to produce resources.}
\footnotetext[2]{Clay can be dug up, and used to make bricks (by smelting clay in a furnace) or clay blocks}
\footnotetext[3]{Wheat can be grown by using seeds on farmland (found by destroying tall grass or wheat). It can form a Hay Block.}
\footnotetext[4]{It must be planted on some specific blocks such as grass, dirt, or sand, which are directly adjacent to water, waterlogged block, or frosted ice, or on top of another sugar cane block.}
\footnotetext[5]{Farmland allows a player to plant seeds, requiring water close to stay fertile.}
\footnotetext[6]{Bone meal can work as a fertilizer; using bone meal on a grass block will grow tall grass, yellow and red flowers on grass blocks around it, allowing the player to harvest large amounts of grass and flowers in a short time.}
\footnotetext[7]{Light-emitting block which causes fire damage when stepped on}

\subsection{Redstone Components}
\begin{center}
\begin{table}
\caption {Redstone Power Component} \label{tab:power} 
\footnotesize
\begin{tabular}{ |p{3cm}|p{11cm}| }
\hline
Redstone Block & This power source has the particularity of being always active. \\ 
\hline
Redstone Torch (/unlit) & provides a continuous pulse and is definitely deactivated (after one tick) if attached to a block of redstone or another powdered block. Often used in redstone clocks.  \\
\hline
Lever & It sends a continuous pulse, which allows activating resp. stop a mechanism if pulled up, resp. down.\\
\hline
Observer & It emits a redstone signal when the block or fluid it faces experience a change.\\
\hline
Wooden/Stone Button & It enables to activate mechanisms for a short time as it sends a small pulse (1s) through solid blocks before being turned off again.\\
\hline
Pressure Plate (e.g. Wooden/Stone, Light/Heavy, etc.) &  It sends a continuous pulse when players or mobs stand on it. The wooden pressure plate can also be activated by dropping an item on it, or by shooting an arrow on it for instance. \\
\hline
Detector Rail & Similarly to pressure plates, when a minecart rides over it, a pulse is sent through the sides of the rails, as long as it remains on it.\\
\hline
Tripwire Hook &  If part of a 'tripwire circuit', it sends a signal when some entities intersect the tripwire line and are deactivated when nothing is in contact with it anymore; it can be used to detect entities over a large area. \\
\hline
Daylight Detector (possibly Inverted) & It outputs a redstone signal based on sunlight.\\
\hline
\end{tabular}
\end{table}
\end{center}

\begin{center}
\begin{table}
\caption {Redstone Transmission Component} \label{tab:transmission} 
\footnotesize
\begin{tabular}{ |p{3cm}|p{11cm}| }
\hline
Powered/Unpowered Repeater & It repeats any pulse it receives, with a delay of 0.1 seconds (up to 0.4). It enables us to overcome the 15 block limit of red stone circuits.\\ 
\hline
Powered/Unpowered Comparator & If its sides are not powered, it outputs the same signal strength as its rear input; whereas in comparison mode, it compares its rear input to its two side inputs and turns off when either side input is greater than the rear input (else it outputs the same signal as its rear input).\\
 \hline
Redstone Wire & Placing Redstone Dust on a block makes it a conductive Redstone Wire: it allows any pulse to pass through it. The maximum range of a pulse is 15 blocks, but repeaters can overcome this limit. Redstone Dust can be mined from Redstone Ore.\\
\hline
Tripwire (/Hook) & Serves to build a tripwire circuit which is made of: a block with a tripwire hook attached to it, some aligned tripwire blocks (1 to 40), and a second tripwire hook attached to another block, with both tripwire hooks facing the tripwire.\\ 
\hline
\end{tabular}
\end{table}
\end{center}

\begin{center}
\begin{table}
\caption {Redstone Mechanism Component} \label{tab:mechanism} 
\footnotesize
\begin{tabular}{ |p{3cm}|p{11cm}| }
\hline
Redstone Lamp & Requires a power source to be lit, and can be toggled on and off.\\ 
\hline
Piston (/Extension, /Head) & Once activated, a piston pushes many blocks in front of its wooden plate. Depending on the block types, these blocks may be pushed, broken by the piston, or even blocking the piston.\\ 
\hline
Sticky piston & These pistons are pushing a block when activated and pulling back the attached block when deactivated. As the piston block, it is an essential block type to automate some processes or for moving machines\\
\hline
Door (e.g. Iron, Wooden, or Trapdoor) & Upon activation, the doors may open.\\
\hline
Dropper/Dispenser & Upon activation, the dropper will eject one item from its inventory after waiting for 2 ticks whereas the dispenser may either fire or drop one item, depending on the item type. Multiple items can be ejected by repeatedly activating the dropper with a clock circuit. \\
\hline
Hopper  & Upon activation, the hopper gets locked; it can be used to catch item entities. \\
\hline
Beacon & Upon activation, it projects a light beam into the sky and gives nearby players status effects such as Speed, Jump Boost, Haste, Regeneration, Resistance, or Strength.\\
\hline
TNT & This explosive block --destructing components-- can be triggered by hitting it or via redtone. \\
\hline
Activator Rail & If powered, it can cause any minecart that runs over them to drop off any player or material that is being transported.\\
\hline
Noteblock & Once activated, it plays a note. Songs can be created with elaborate red stone circuits.\\
\hline
Command block & Command blocks can execute commands when activated, as a sort of pseudo-programming language. They may have different modes (e.g. conditional) and can be chained/repeated. \\
\hline

\end{tabular}
\end{table}
\end{center}

\newpage

\begin{lstlisting}[language=python,style=protobuf, breaklines=false, caption={Example use of the EvoCraft Python client to implement a simple evolutionary algorithm used in Section~\ref{sec:automated_evolution}.}, label=listing:python-evol, captionpos=b]
def evolution(gens=1000, pop_num=200, mutation_p=0.1, parent_cut_r=0.1):
    population_coordinates = [(start_coord[0],
                               start_coord[1],
                               start_coord[2] + (i * 20))
                              for i in range(pop_num)]
    population = [generate_individual(c) for c in population_coordinates]
    block_buffer = BlockBuffer()
    for generation in range(gens):
        if generation == 0 or generation == 1:
            time.sleep(50)

        # Clear the area
        client.fillCube(FillCubeRequest(
            cube=Cube(
            min=Point(x=slots_start[0], y=slots_start[1], z=slots_start[2]),
            max=Point(x=slots_end[0], y=slots_end[1], z=slots_end[2])
            ),
            type=AIR
        ))
        # Show population
        client.spawnBlocks(Blocks([ind.get_blocks()
                                   for ind in population]))
        print(f"Generation --> {generation}")
        get_ind_blocks = lambda i: client.readCube(min=i.slot_coordinates()[0],
                                                   max=i.slot_coordinates()[1])
        population_blocks = list(map(evaluate, population))
        evaluations = map(evaluate, population_blocks)
        pop_eval_zipped = zip(population, evaluations)
        pop_eval_zipped = \
            sorted(pop_eval_zipped, key=itemgetter(1), reverse=False)
        sorted_population, _ = map(list, zip(*pop_eval_zipped))
        parent_cuttoff_idx = int(parent_cut_r * pop_num)
        parents = sorted_population[:parent_cuttoff_idx]

        next_generation = [clone_individual(parents[0])]
        for _ in range(pop_num - 1):
            p1 = parents[randint(0, len(parents))]
            p2 = parents[randint(0, len(parents))]
            c = crossover(p1, p2)

            if uniform(0, 1) < mutation_p:
                c = mutation(c[0])
            next_generation.append(c)
        population = next_generation
\end{lstlisting}


\end{document}